\documentclass[10pt,twocolumn,letterpaper]{article}

\usepackage{cvpr}
\usepackage[utf8]{inputenc}
\usepackage{times}
\usepackage{epsfig}
\usepackage{graphicx}
\usepackage{amsmath,mathtools}
\usepackage{amssymb}
\usepackage{amsthm}
\usepackage{bm}
\usepackage{enumitem}
\usepackage{mathtools}
\usepackage{multirow}
\usepackage[htt]{hyphenat}
\usepackage{fontawesome}
\usepackage{pifont}
\usepackage{color}
\usepackage{float}
\usepackage{xcolor}
\usepackage{subfig}
\usepackage[algo2e,inoutnumbered,linesnumbered,algoruled,vlined]{algorithm2e}
\usepackage{algpseudocode}
\usepackage{booktabs}
\usepackage{tabularx}
\usepackage{bm}
\usepackage{url}
\usepackage{bbm}
\usepackage{leftidx}
\usepackage{dcolumn}
\usepackage{enumitem}
\usepackage{gensymb}
\usepackage{tikz}
\usetikzlibrary{patterns}

\usepackage[pagebackref=true,breaklinks=true,colorlinks,bookmarks=false]{hyperref}

\newcolumntype{d}[1]{D{.}{.}{#1}}
\newcolumntype{B}[3]{>{\boldmath\DC@{#1}{#2}{#3}}c<{\DC@end}}

\definecolor{tikz_gray}{RGB}{191,191,191}
\definecolor{tikz_lblue}{RGB}{93,138,210}
\definecolor{tikz_dblue}{RGB}{46,78,124}
\definecolor{tikz_lviolet}{RGB}{161,125,173}
\definecolor{tikz_dviolet}{RGB}{148,55,255}
\definecolor{tikz_rose}{RGB}{250,150,150}
\definecolor{tikz_pink}{RGB}{214,19,115}
\definecolor{tikz_lpink}{RGB}{255,138,216}

\newcommand{\overbar}[1]{\mkern 1.5mu\overline{\mkern-1.5mu#1\mkern-1.5mu}\mkern 1.5mu}

\newcommand{\parahead}[1]{\noindent\textbf{#1}:\ }

\newcommand{\acro}{\textsc{Predator}}
\newcommand{\acroexplain}{\textbf{p}oint-cloud \textbf{re}gistration with \textbf{d}eep \textbf{at}tention to the \textbf{o}verlap \textbf{r}egion}

\usepackage{authblk}
\makeatletter
\newcommand{\printfnsymbol}[1]{%
  \textsuperscript{\@fnsymbol{#1}}%
}
\makeatother

\cvprfinalcopy %

\def\cvprPaperID{1139} %

\ifcvprfinal\fi

\usepackage{cleveref}
\crefname{section}{\S}{\S\S}
\crefname{subsection}{\S}{\S\S}

\Crefname{assumption}{\textbf{H}\hspace{-3pt}}{\textbf{H}\hspace{-3pt}}
\crefname{assumption}{\textbf{H}}{\textbf{H}}

\begin{document}

\title{\acro: Registration of 3D Point Clouds with Low Overlap}

\author{\vspace{-20pt}Shengyu Huang$^{*}$ \quad Zan Gojcic$^{*}$ \quad Mikhail Usvyatsov \quad Andreas Wieser \quad Konrad Schindler \\ETH Zurich \\\url{overlappredator.github.io}}

\maketitle
\begin{abstract}
We introduce \acro, a model for pairwise \acroexplain. Different from previous work, our model is specifically designed to handle (also) point-cloud pairs with low overlap.  Its key novelty is an  overlap-attention block for early information exchange between the latent encodings of the two point clouds. In this way the subsequent decoding of the latent representations into per-point features is conditioned on the respective other point cloud, and thus can predict which points are not only salient, but also lie in the overlap region between the two point clouds. The ability to focus on points that are relevant for matching greatly improves performance: \acro\ raises the rate of successful registrations by more than 15 percent points in the low-overlap scenario, and also sets a new state of the art for the \emph{3DMatch} benchmark with 90.6\% registration recall. [\href{https://github.com/ShengyuH/OverlapPredator}{Code release}]  
\end{abstract}
\section{Introduction}
\label{sec:intro}
{\let\thefootnote\relax\footnote{{$^*$First two authors contributed equally to this work.}}}

Recent work has made substantial progress in fully automatic, 3D feature-based point cloud registration. At first glance, benchmarks like \textit{3DMatch}~\cite{zeng20163dmatch} appear to be saturated, with multiple state-of-the-art (SoTA) methods~\cite{gojcic20193DSmoothNet,Choy2019FCGF,bai2020d3feat} reaching nearly 95\% feature matching recall and successfully registering $>$80\% of all scan pairs.
One may get the impression that the registration problem is solved---but this is actually not the case. We argue that the high success rates are a consequence of lenient evaluation protocols. We have been making our task too easy:
existing literature and benchmarks~\cite{choi2015robust, zeng20163dmatch, khoury2017CGF} consider only pairs of point clouds with $\geq$30\% overlap to measure performance.
Yet, the low-overlap regime is very relevant for practical applications. On the one hand, it may be difficult to ensure high overlap, for instance when moving along narrow corridors, or when closing loops in the presence of occlusions (densely built-up areas, forest, etc.). On the other hand, data acquisition is often costly, so practitioners aim for a low number of scans with only the necessary overlap~\cite{yang2019extreme,yang2020extreme}. 

\begin{figure}[t]
    \centering
    \includegraphics[width=1.0\columnwidth]{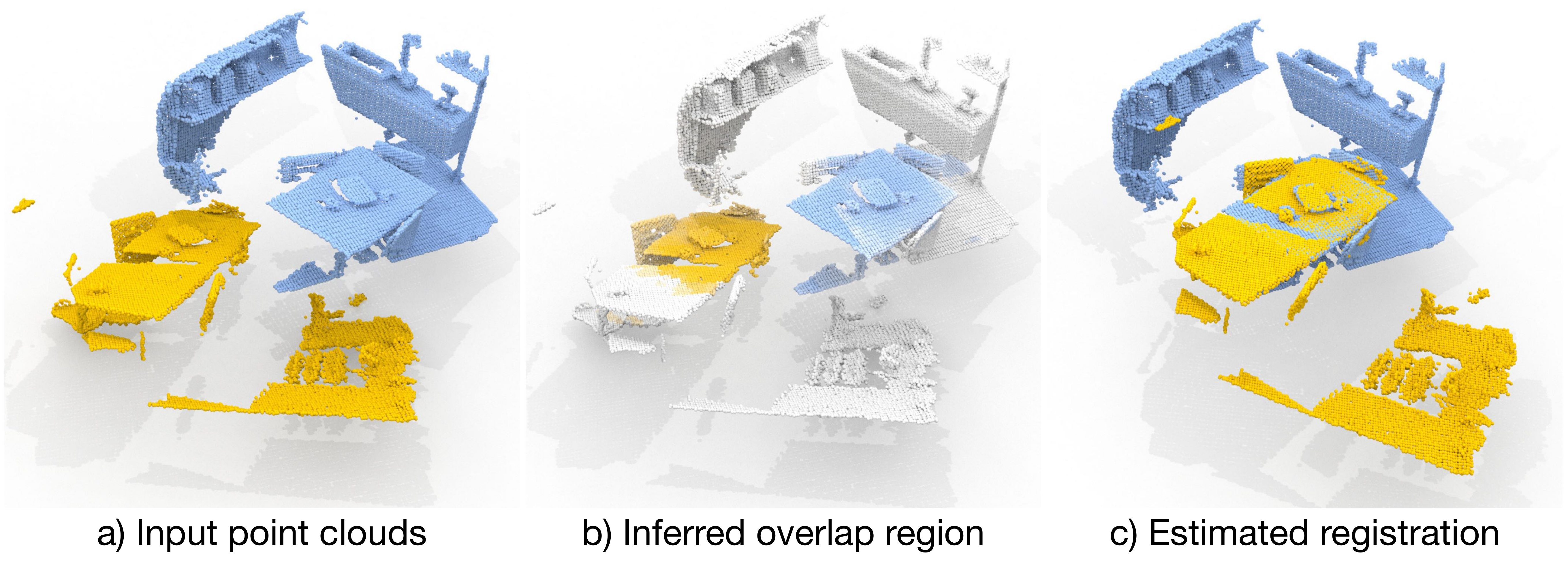}
    \caption{\acro\ is designed to focus attention on the overlap region, and to prefer salient points in that region, so as to enable robust registration in spite of low overlap.}
    \label{fig:teaser_image}
\end{figure}
\begin{figure}[t]
    \centering

    \includegraphics[width=\columnwidth]{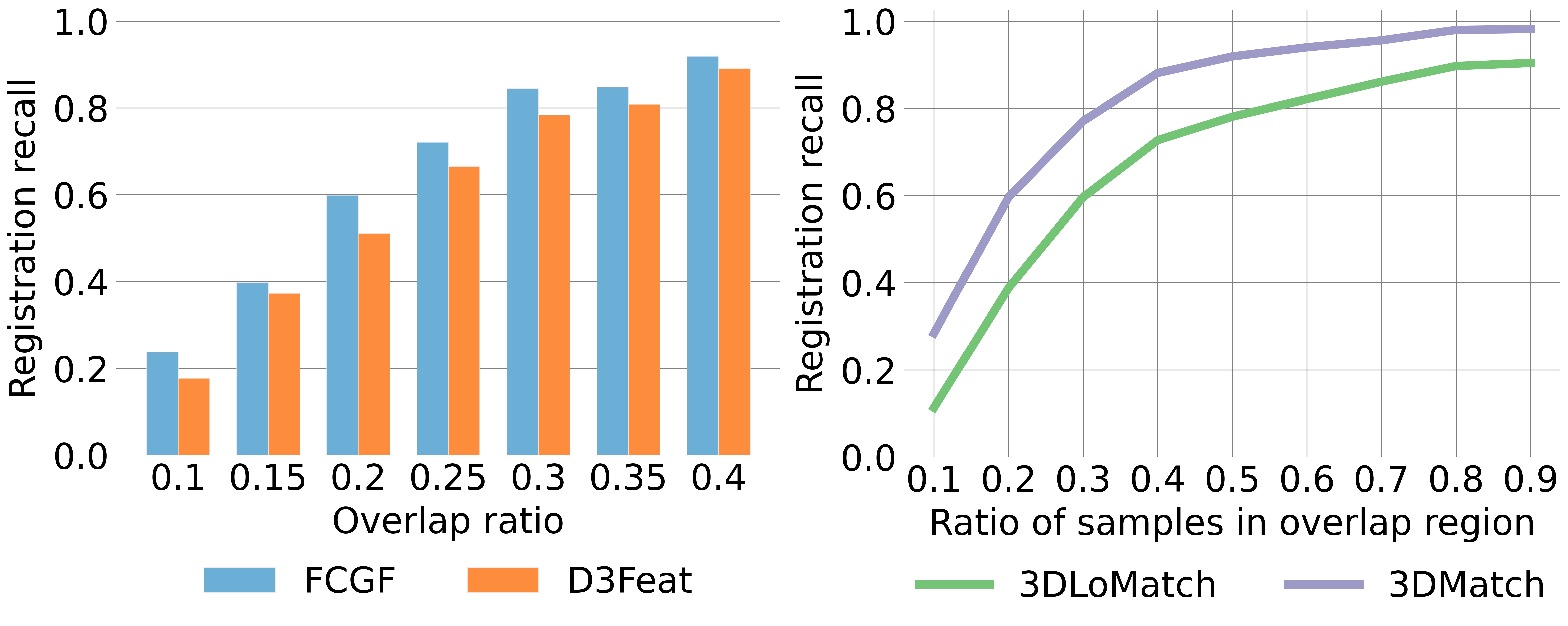}\vspace{-2mm}
    \caption{Registration with SoTA methods deteriorates rapidly for pairs with $<$30\% overlap (\textit{left}). By increasing the fraction of points sampled in the overlap region, many failures can be avoided as shown here for FCGF~\cite{Choy2019FCGF} (\textit{right}).} 
    \label{fig:motivation}
    \vspace{-\baselineskip}
\end{figure}

Driven by the evaluation protocol, the high-overlap scenario became the focus of research, whereas the more challenging low-overlap examples were largely neglected~(\cf~Fig.~\ref{fig:teaser_image}).
Consequently, the registration performance of even the best known methods deteriorates rapidly when the overlap between the two point clouds falls below 30\%, see Fig.~\ref{fig:motivation}.
Human operators, in contrast, can still register such low overlap point clouds without much effort.%

This discrepancy is the starting point of the present work. %
To %
study its reasons, we have constructed a low-overlap dataset \textit{3DLoMatch} from scans of the popular \textit{3DMatch} benchmark, and have analysed the individual modules/steps of the registration pipeline (Fig.~\ref{fig:motivation}).
It turns out that the effective receptive field of fully convolutional feature point descriptors~\cite{Choy2019FCGF,bai2020d3feat} is local enough and the descriptors are hardly corrupted by non-overlapping parts of the scans. Rather than coming up with yet another way to learn better descriptors, the key to registering low overlap point clouds is \emph{learning where to sample feature points}. A large performance boost can be achieved if the feature points are predominantly sampled from the overlapping portions of the scans~(Fig.~\ref{fig:motivation}, right).

We follow this path and introduce \acro,
a neural architecture for pairwise 3D point cloud registration that learns to detect the overlap region between two unregistered scans, and to focus on that region when sampling feature points.
The main contributions of our work are:
\begin{itemize}[leftmargin=15pt,topsep=4pt]
\setlength{\itemsep}{0pt}
\setlength{\parskip}{2pt}
    \item an analysis why existing registration pipelines break down in the low-overlap regime
    \item a novel \emph{overlap attention} block that allows for early information exchange between the two point clouds and focuses the subsequent steps on the overlap region
    \item a scheme to refine the feature point descriptors, by conditioning them also on the respective other point cloud
    \item a novel loss function to train \emph{matchability} scores, which help to sample better and more repeatable interest points
\end{itemize}
Moreover, we make available the \textit{3DLoMatch} dataset, containing the previously ignored scan pairs of \textit{3DMatch} that have low (10-30\%) overlap.
In our experiments, \acro\ greatly outperforms existing methods in the low-overlap regime, increasing registration recall by \textgreater15 percent points.
It also sets a new state of the art on the \textit{3DMatch} benchmark, reaching a registration recall of \textgreater90\%.

\section{Related work}
\label{sec:related_work}
\begin{figure*}[t]
    \centering
    \includegraphics[width=0.9\textwidth]{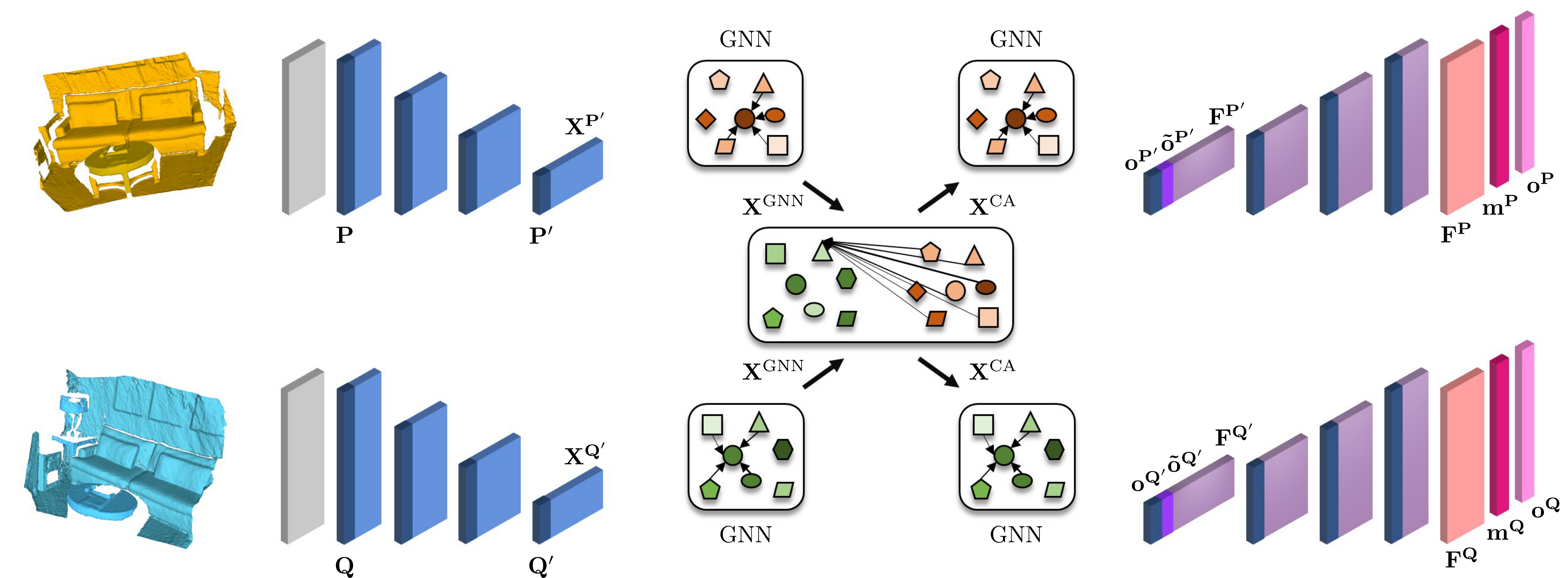}
    \caption{Network architecture of \acro. \textcolor{gray}{Voxel-gridded} point clouds $\mathbf{P}$ and $\mathbf{Q}$ are fed to the encoder, which extracts the superpoints $\mathbf{P}'$ and $\mathbf{Q}'$ and their latent features $\mathbf{X}^{\mathbf{P}'}$, $\mathbf{X}^{\mathbf{Q}'}$. The overlap-attention module updates the features with co-contextual information in a series of self- (GNN) and cross-attention (CA) blocks, and projects them to overlap  $\mathbf{o}^{\mathbf{P}'}$, $\mathbf{o}^{\mathbf{Q}'}$ and cross-overlap $\tilde{\mathbf{o}}^{\mathbf{P}'}$, $\tilde{\mathbf{o}}^{\mathbf{Q}'}$ scores. Finally, the decoder transforms the conditioned features and overlap scores to per-point feature descriptors $\mathbf{F}^\mathbf{P}$, $\mathbf{F}^\mathbf{Q}$, overlap scores $\mathbf{o}^\mathbf{P}$, $\mathbf{o}^\mathbf{Q}$, and matchability scores $\mathbf{m}^\mathbf{P}$, $\mathbf{m}^\mathbf{Q}$.}
    \label{fig:network_arch}
    \vspace{-\baselineskip}
\end{figure*}

We start this related-work section by reviewing the individual components of the traditional point cloud registration pipelines, before proceeding to newer, end-to-end point-cloud registration algorithms. Finally, we briefly cover recent advances in using contextual information to guide and robustify feature extraction and matching.

\parahead{Local 3D feature descriptors} 
Early local descriptors for point clouds~\cite{johnson1999, rusu2008PFH, rusu2009FPFH, tombari2010SHOT, tombari2010USC} aimed to characterise the local geometry by using hand-crafted features. While often lacking robustness against clutter and occlusions, they have long been a default choice for  downstream tasks because they naturally generalise across datasets~\cite{guo2014performanceEvaluation}. In the last years, learned 3D feature descriptors have taken 
over and now routinely outperform their hand-crafted counterparts.

The pioneering 3DMatch method~\cite{zeng20163dmatch} is based on a Siamese 3D CNN that extracts local feature descriptors from a signed distance function embedding.
Others~\cite{khoury2017CGF, gojcic2018learned} first extract hand-crafted features, then map them to a compact representation using multi-layer perceptrons. PPFNet~\cite{deng2018ppfnet}, and its self-supervised version PPF-FoldNet~\cite{Deng2018PPFFoldNetUL}, combine point pair features with a PointNet~\cite{qi2017pointnet} architecture to extract descriptors that are aware of the global context. To alleviate artefacts caused by noise and voxelisation,~\cite{gojcic20193DSmoothNet} proposed to use a smoothed density voxel grid as input to a 3D CNN. These early works achieved strong performance, but still operate on individual local patches, which greatly increases the computational cost and limits the receptive field to a predefined size.

Fully convolutional architectures~\cite{long2015fully} that enable dense feature computation over the whole input in a single forward pass~\cite{detone2018superpoint, dusmanu2019d2Net, revaud2019r2d2} have been adopted to design faster 3D feature descriptors. Building on sparse convolutions~\cite{choy2019Minkowski}, FCGF~\cite{Choy2019FCGF} achieves
a performance similar to the best patch-based descriptors~\cite{gojcic20193DSmoothNet}, while being orders of magnitude faster. D3Feat~\cite{bai2020d3feat} complements
a fully convolutional feature descriptor with an salient point detector.

\parahead{Interest point sampling} 
The classic principle to sample salient rather than random  points
has also found its way into learned 2D~\cite{detone2018superpoint, dusmanu2019d2Net,revaud2019r2d2, wiles2020d2d} and 3D~\cite{yew20183dfeat, bai2020d3feat,lu2020rskdd} local feature extraction.
All these methods implicitly assume that the saliency of a point fully determines its utility for downstream tasks. Here, we take a step back and argue that, while saliency is desirable for an interest point, it is not sufficient on its own. Indeed, in order to contribute to registration a point should not only be salient, but must also lie in the region where the two point clouds overlap---an essential property that, surprisingly, has largely been neglected thus far.

\parahead{Deep point-cloud registration} Instead of combining learned feature descriptors with some
off-the-shelf robust optimization at inference time, a parallel stream of work aims to embed the differentiable pose estimation into the learning pipeline. PointNetLK~\cite{aoki2019pointnetlk} combines a PointNet-based global feature descriptor~\cite{qi2017pointnet} with a Lucas/Kanade-like optimization algorithm~\cite{lucas1981LK} and estimates the relative transformation in an iterative fashion.
DCP~\cite{wang2019dcp} use a DGCNN network~\cite{wang2019dynamic} to extract local features and computes soft correspondences before using the Kabsch algorithm to estimate the transformation parameters. To relax the need for strict one-to-one correspondence, DCP was later extended to PRNet~\cite{wang2019prnet}, which includes a \textit{keypoint} detection step and allows for partial correspondence. Instead of simply using soft correspondences, ~\cite{yew2020rpm} %
update the similarity matrix with a differentiable Sinkhorn layer~\cite{sinkhorn1964relationship}. Similar to other methods, the weighted Kabsch algorithm\cite{4767965} is used to estimate the transformation parameters. Finally, ~\cite{gojcic2020learning,choy2020deep,pais20203dregnet}
complement a learned feature descriptor with an outlier filtering network, which infers the correspondence weights for later use in the weighted Kabsch algorithm.

\parahead{Contextual information} In the traditional pipeline, feature extraction is done independently per point cloud. Information %
is only communicated when computing pairwise similarities, although aggregating contextual information
at an earlier stage could provide additional cues to robustify the descriptors and guide the matching step.

In 2D feature learning, D2D-Net~\cite{wiles2020d2d} use an attention mechanism in the bottleneck of an encoder-decoder scheme to aggregate the contextual information, which is later used to condition the output of the decoder on the second image.
SuperGlue~\cite{sarlin2020superglue} infuses the contextual information into the learned descriptors with a whole series of self- and cross-attention layers, built upon the message-passing GNN~\cite{kipf2016semi}. 
Early information mixing was previously also explored in the field of deep point cloud registration, where~\cite{wang2019dcp, wang2019prnet} use a transformer module to extract task-specific 3D features that are reinforced with contextual information.

\section{Method}
\label{sec:method}

\acro\ is a two-stream encoder-decoder network. Our default implementation uses residual blocks with KPConv-style point convolutions~\cite{thomas2019kpconv}, but the architecture is agnostic w.r.t.\ the backbone and can also be implemented with other formulations of 3D convolutions, such as for instance sparse voxel convolutions~\cite{choy2019Minkowski} (\cf Appendix). 
As illustrated in Fig.~\ref{fig:network_arch}, the architecture of \acro\ can be decomposed into three main modules:
\begin{enumerate}[topsep=2pt,itemsep=0pt,partopsep=1ex,parsep=2pt,leftmargin=15pt]
\item encoding of the two point clouds into smaller sets of superpoints and associated latent feature encodings, with shared weights (Sec.~\ref{sec:method_encoder});
\item the overlap attention module (in the bottleneck) that extracts co-contextual information between the feature encodings of the two point clouds, and assigns each superpoint two overlap scores that quantify how likely the superpoint itself and its soft-correspondence are located in the overlap between the two inputs (Sec.~\ref{sec:method_overlap_attention}); 
\item decoding of the mutually conditioned bottleneck representations to point-wise descriptors as well as refined per-point overlap and matchability scores (Sec.~\ref{sec:method_decoder}).
\end{enumerate}
Before diving into each component we lay out the basic problem setting and notation in Sec.~\ref{sec:method_notation}. %
\subsection{Problem setting}
\label{sec:method_notation}

Consider two point clouds $\mathbf{P}= \{\mathbf{p}_i \in \mathbb{R}^3| i = 1..N\}$, and $\mathbf{Q}= \{\mathbf{q}_i \in \mathbb{R}^3| i = 1..M\}$.
Our goal is to recover a rigid transformation $\mathbf{T}_\mathbf{P}^\mathbf{Q}$ with parameters $\mathbf{R} \in SO(3)$ and $\mathbf{t} \in \mathbb{R}^3$ that aligns %
$\mathbf{P}$ to $\mathbf{Q}$.
By a slight abuse of notation we use the same symbols for sets of points and for their corresponding matrices $\mathbf{P}\in\mathbb{R}^{N\times 3}$ and $\mathbf{Q}\in\mathbb{R}^{M\times 3}$.

\begin{figure}[t]
    \centering
    \includegraphics[width=\columnwidth]{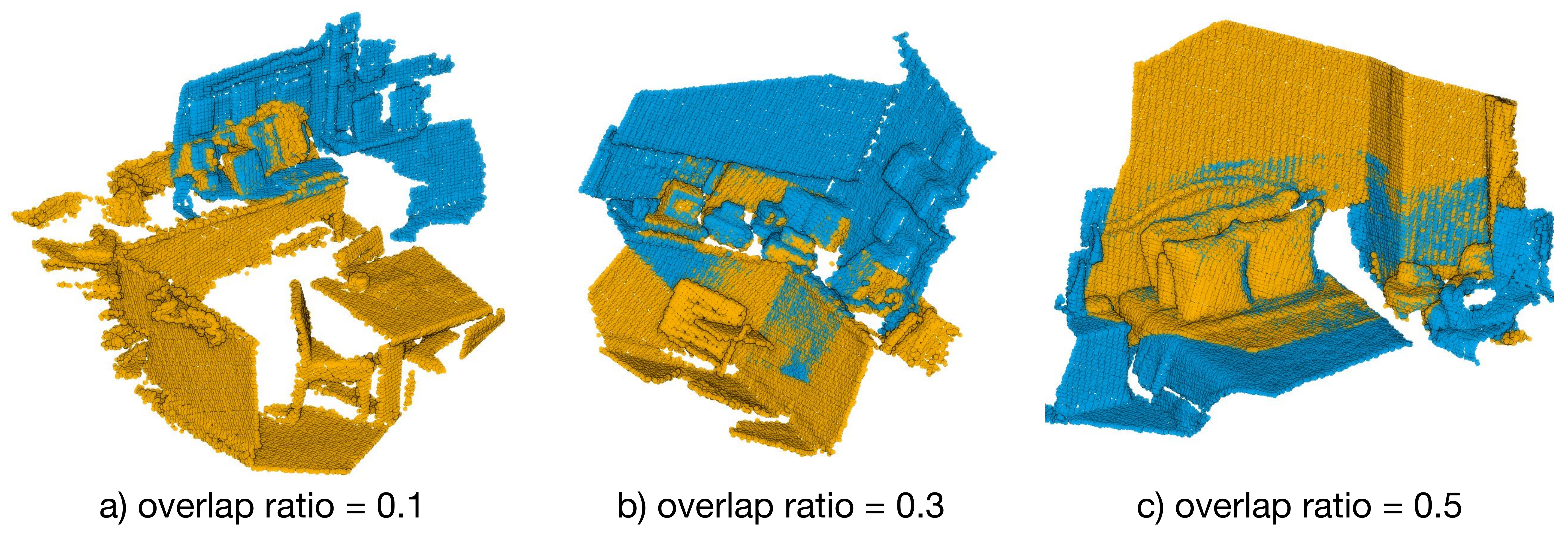}\vspace{-2mm}
    \caption{Fragments with different overlap ratios. Overlap is computed relative to the source fragment (orange).}
    \label{fig:demo_overlap}
    \vspace{-\baselineskip}
\end{figure}

Obviously $\mathbf{T}_\mathbf{P}^\mathbf{Q}$ can only ever be determined from the data if $\mathbf{P}$ and $\mathbf{Q}$ have sufficient overlap, meaning that after applying the ground truth transformation $\overbar{\mathbf{T}}_\mathbf{P}^\mathbf{Q}$ the overlap ratio
\begin{equation}
 \frac{1}{N}\big|\big\{\|(\overbar{\mathbf{T}}_\mathbf{P}^ \mathbf{Q}(\mathbf{p}_i)-\mathsf{NN}(\overbar{\mathbf{T}}_\mathbf{P}^\mathbf{Q}(\mathbf{p}_i),\mathbf{Q})\|_2%
 \leq v\big\}\big|>\tau\;,
    \label{eq:overlap_ratio}
\end{equation}
where $\mathsf{NN}$ denotes the nearest-neighbour operator w.r.t.\ its second argument, $\|\!\cdot\!\|_2$ is the Euclidean norm, $|\!\cdot\!|$ is the set cardinality, and $v$ is a tolerance that depends on the point density.%
\footnote{For efficiency, $v$ is in practice determined after voxel-grid down-sampling of the two point clouds.} %
Contrary to previous work~\cite{zeng20163dmatch,khoury2017CGF}, where the threshold to even attempt the alignment is typically $\tau\!>\!0.3$, we are interested in low-overlap point clouds with $\tau\!>\!0.1$. Fragments with different overlap ratios are shown in Fig.~\ref{fig:demo_overlap}.

\subsection{Encoder}
\label{sec:method_encoder}
We follow~\cite{thomas2019kpconv} and first down-sample raw point clouds with a voxel-grid filter of size $V$, such that
$\mathbf{P}$ and $\mathbf{Q}$ have reasonably uniform point density.
In the shared encoder, a series of ResNet-like blocks and strided convolutions aggregate the raw points into \textit{superpoints} $\mathbf{P}'\in\mathbb{R}^{N'\times 3}$ and 
$\mathbf{Q}'\in\mathbb{R}^{M'\times 3}$ with associated features $\mathbf{X}^{\mathbf{P}'} \in \mathbb{R}^{N{'} \times b}$ and $\mathbf{X}^{\mathbf{Q}'} \in \mathbb{R}^{M' \times b}$.
Note that superpoints correspond to a fixed receptive field, so their number depends on the spatial extent of the input point cloud and may be different for the two inputs.

\subsection{Overlap attention module}
\label{sec:method_overlap_attention}
So far, the features $\mathbf{X}^{\mathbf{P}'}$, $\mathbf{X}^{\mathbf{Q}'}$ in the bottleneck encode the geometry and context of the two point clouds. But $\mathbf{X}^{\mathbf{P}'}$ has no knowledge of point cloud $\mathbf{Q}$ and vice versa.
In order to reason about their respective overlap regions, some cross-talk is necessary.
We argue that it makes sense to add that cross-talk at the level of superpoints in the bottleneck, just like a human operator will first get a rough overview of the overall shape to determine likely overlap regions, and only after that identifies precise feature points in those regions.

\parahead{Graph convolutional neural network}
Before connecting the two feature encodings, we first further aggregate and strengthen their contextual relations individually with a graph neural network (GNN)~\cite{wang2019dynamic}.
In the following, we describe the GNN for point cloud $\mathbf{P}'$.
The GNN for $\mathbf{Q}'$ is the same.
First, the superpoints in $\mathbf{P}'$ are linked into a graph in Euclidean space with the $k$-NN method.
Let ${\mathbf{x}}_i \in \mathbb{R}^{b}$ denote the feature encoding of superpoint $\mathbf{p}'_i$, and $(i,j) \in \mathcal{E}$ the graph edge between superpoints $\mathbf{p}_i'$ and $\mathbf{p}_j'$.
The encoder features are then iteratively updated as
\begin{equation}
    \leftidx{^{(k+1)}}{\mathbf{x}}_i = \max_{(i,j)\in\mathcal{E}} h_\theta\big( \mathrm{cat}[\leftidx{^{(k)}}{\mathbf{x}}_i, \leftidx{^{(k)}}{\mathbf{x}}_j - \leftidx{^{(k)}}{\mathbf{x}}_i]\big)\;,
\end{equation}
where $h_\theta(\cdot)$ denotes a linear layer followed by instance normalization~\cite{ulyanov2016instance} and a LeakyReLU activation~\cite{maas2013rectifier}, $\max(\cdot)$ denotes  element-/channel-wise max-pooling, and $\mathrm{cat}[\cdot,\cdot]$ means concatenation. This update is performed twice with separate (not shared) parameters $\theta$, and the final GNN features ${\mathbf{x}}_i^{\mathrm{GNN}} \in \mathbb{R}^{d_b}$ are obtained as
\begin{equation}
    {\mathbf{x}}_i^{\mathrm{GNN}} =  h_\theta(\text{cat}[\leftidx{^{(0)}}{\mathbf{x}}_i, \leftidx{^{(1)}}{\mathbf{x}}_i,
    \leftidx{^{(2)}}{\mathbf{x}}_i])\;.
\end{equation}

\parahead{Cross-attention block} 
Knowledge about potential overlap regions can only be gained by mixing information about both point clouds. To this end we adopt a cross-attention block~\cite{sarlin2020superglue} based on the message passing formulation~\cite{gilmer2017neural}.
First, each superpoint in $\mathbf{P}'$ is connected to all superpoints in $\mathbf{Q}'$ to form a bipartite graph.
Inspired by the Transformer architecture~\cite{vaswani2017attention}, vector-valued queries $\mathbf{s}_i\!\in\!\mathbb{R}^{b}$ are used to retrieve the values $\mathbf{v}_j\!\in\!\mathbb{R}^{b}$ of other superpoints based on their keys
$\mathbf{k}_j\!\in\!\mathbb{R}^{b}$, where 
\begin{equation}
    \mathbf{k}_j = \mathbf{W}_k\mathbf{x}_j^{\mathrm{GNN}}\quad
    \mathbf{v}_j = \mathbf{W}_v\mathbf{x}_j^{\mathrm{GNN}}\quad
    \mathbf{s}_i = \mathbf{W}_s\mathbf{x}_i^{\mathrm{GNN}}\\
\end{equation}
and $\mathbf{W}_k$, $\mathbf{W}_v$, and $\mathbf{W}_s$ are learnable weight matrices.
The messages are computed as weighted averages of the values,
\begin{equation}
    \mathbf{m}_{i\leftarrow} = \sum_{j:(i,j)\in\mathcal{E}}a_{ij}\mathbf{v}_j\;,
\end{equation}
with attention weights $a_{ij} = \text{softmax}(\mathbf{s}^{T}_i\mathbf{k}_j / \sqrt{b})$~\cite{sarlin2020superglue}.
I.e., to update a superpoint $\mathbf{p}_i'$ one combines that point's query with the keys and values of all superpoints $\mathbf{q}_j'$.
In line with the literature, in practice we use a multi-attention layer with four parallel attention heads
~\cite{vaswani2017attention}.
The co-contextual features are computed as
\begin{equation}
    \mathbf{x}_i^{\text{CA}} = \mathbf{x}_i^{\mathrm{GNN}} + \mathrm{MLP}(\mathrm{cat}[\mathbf{s}_i, \mathbf{m}_{i\leftarrow}])\;,
\end{equation}
with $\mathrm{MLP}(\cdot)$ denoting a three-layer fully connected network with instance normalization~\cite{ulyanov2016instance} and ReLU~\cite{nair2010rectified} activations after the first two layers.
The same cross-attention block is also applied in reverse direction, so that information flows in both directions, $\mathbf{P}'\!\rightarrow\!\mathbf{Q}'$ and $\mathbf{Q}'\!\rightarrow\!\mathbf{P}'$.

\parahead{Overlap scores of the bottleneck points}
The above update with co-contextual information is done for each superpoint in isolation, without considering the local context within each point cloud.
We therefore, explicitly update the local context after the cross-attention block using another GNN that has the same architecture and underlying graph (within-point cloud links) as above, but separate parameters $\theta$.
This yields the final latent feature encodings  $\mathbf{F}^{\mathbf{P}'}\!\!\in\!\mathbb{R}^{N'\!\times b}$ and $\mathbf{F}^{\mathbf{Q}'}\!\!\in\!\mathbb{R}^{M'\!\times b}$, which are now conditioned on the features of the respective other point cloud.
Those features are linearly projected to overlap scores $\mathbf{o}^{\mathbf{P}'}\!\!\in\!\mathbb{R}^{N'}$ and $\mathbf{o}^{\mathbf{Q}'}\!\!\in\!\mathbb{R}^{M'}$, which can be interpreted as probabilities that a certain superpoint lies in the overlap region.
Additionally, one can compute \emph{soft correspondences} between superpoints and from the correspondence weights predict the \emph{cross-overlap score} of a superpoint $\mathbf{p}'_i$, i.e., the probability that its
correspondence in $\mathbf{Q}'$ lies in the overlap region:
\begin{equation}
    \tilde{o}_i^{\mathbf{P}'}:= \mathbf{w}_i^{T} \mathbf{o}^{\mathbf{Q}'}, \quad w_{ij} := \mathrm{softmax}\big(\frac{1}{t}\langle\mathbf{f}^{\mathbf{P}'}_i, \mathbf{f}^{\mathbf{Q}'}_j\rangle\big)\;,
    \label{eq:softass}
\end{equation}
where $\langle \cdot,\cdot \rangle$ is the inner product, and $t$ is the temperature parameter that controls the soft assignment. In the limit $t\!\rightarrow\!0$, Eq.~\eqref{eq:softass} converges to hard nearest-neighbour assignment.

\subsection{Decoder}
\label{sec:method_decoder}
Our decoder starts from conditioned features $\mathbf{F}^{\mathbf{P}'}$, concatenates them with the overlap scores $\mathbf{o}^{\mathbf{P}'}$, $\tilde{\mathbf{o}}^{\mathbf{P}'}$, and outputs per-point feature descriptors $\mathbf{F}^{\mathbf{P}}\!\!\in\!\mathbb{R}^{N\times32}$ and refined per-point overlap and matchability scores $\mathbf{o}^{\mathbf{P}},\mathbf{m}^{\mathbf{P}}\!\in \!\mathbb{R}^{N}$.
The matchability can be seen as a "conditional saliency" that quantifies how likely a point is to be matched correctly, given the points (resp.\ features) in the other point cloud $\mathbf{Q}$.

The decoder architecture combines {NN}-upsampling with linear layers, and includes skip connections from the corresponding encoder layers.
We deliberately keep the overlap score and the matchability separate
to disentangle the reasons why a point is a good/bad candidate for matching: in principle a point can be unambiguously matchable but lie outside the overlap region, or it can lie in the overlap but have an ambiguous descriptor.
Empirically, we find that the network learns to predict high matchability mostly for points in the overlap; probably reflecting the fact that the ground truth correspondences used for training, naturally, always lie in the overlap.
For further details about the architecture, please refer to Appendix and the \href{https://github.com/ShengyuH/OverlapPredator}{source code}.

\subsection{Loss function and training}
\label{sec:method_training}
\acro\ is trained end-to-end, using three losses \wrt ground truth correspondences as supervision.

\parahead{Circle loss}
To supervise the point-wise feature descriptors we follow%
\footnote{Added to the repository after publication, not mentioned in the paper.}%
~\cite{bai2020d3feat} and use the circle loss~\cite{sun2020circle}, a variant of the more common triplet loss.
Consider again a pair of overlapping point clouds $\mathbf{P}$ and $\mathbf{Q}$, this time aligned with the ground truth transformation. We start by extracting the points $\mathbf{p}_i\!\in\!\mathbf{P}_p\!\subset\!\mathbf{P}$ that have at least one (possibly multiple) correspondence in $\mathbf{Q}$,
where the set of correspondences $\mathcal{E}_p(\mathbf{p}_i)$  is defined as points in $\mathbf{Q}$ that lie within a radius $r_p$ around $\mathbf{p}_i$. Similarly, all points of $\mathbf{Q}$  outside a (larger) radius
$r_s$ form the set of negatives $\mathcal{E}_n(\mathbf{p}_i)$.
The circle loss is then computed from $n_p$ points sampled randomly from $\mathbf{P}_p$: %
\begin{equation}
\mathcal{L}_c^\mathbf{P} = \frac{1}{n_p}\sum\limits_{i=1}^{n_p} \log \Big [ 1 +\!\! \sum\limits_{j\in\mathcal{E}_p}\!e^{ \beta_p^j(d_i^j - \Delta_p)} \cdot\!
\sum\limits_{k\in\mathcal{E}_n} \!e^{\beta_n^k(\Delta_n - d_i^k)}\Big ],
\end{equation}
where $d_i^j=||\mathbf{f}_{\mathbf{p}_i} - \mathbf{f}_{\mathbf{q}_j} ||_2$ denotes distance in feature space, and $\Delta_n,\Delta_p$ are negative and positive margins, respectively. The weights $\beta_p^j\!=\!\gamma(d_i^j\!-\!\Delta_p)$ and $\beta_n^k\!=\!\gamma(\Delta_n\!-\!d_i^k)$ are determined individually for each positive and negative example, using the empirical margins $\Delta_p\!:=\!0.1$ and $\Delta_n\!:=\!1.4$ with hyper-parameter $\gamma$. The reverse loss  $\mathcal{L}_c^\mathbf{Q}$ is computed in the same way, for a total circle loss $\mathcal{L}_c = \frac{1}{2}(\mathcal{L}_c^\mathbf{P} + \mathcal{L}_c^\mathbf{Q})$.

\parahead{Overlap loss}
The estimation of the overlap probability is cast as binary classification and supervised using the overlap loss $\mathcal{L}_o\!=\!\frac{1}{2} (\mathcal{L}^\mathbf{P}_o + \mathcal{L}^\mathbf{Q}_o)$, where 
\begin{equation}
\mathcal{L}^\mathbf{P}_o = \frac{1}{|\mathbf{P}|} \sum_{i=1}^{|\mathbf{P}|} \overbar{o}_{\mathbf{p}_i}\log(o_{\mathbf{p}_i}) + (1 - \overbar{o}_{\mathbf{p}_i})\log(1 - o_{\mathbf{p}_i}).
\end{equation}
The ground truth label $\overbar{o}_{\mathbf{p}_i}$ of point $\mathbf{p}_i$ is defined as
\begin{equation}
    \overbar{o}_{\mathbf{p}_i} =     
    \begin{cases}
      1, & ||\overbar{\mathbf{T}}_\mathbf{P}^\mathbf{Q}(\mathbf{p}_i) - \mathsf{NN}(\overbar{\mathbf{T}}_\mathbf{P}^\mathbf{Q}(\mathbf{p}_i),\mathbf{Q}) ||_2 < r_o \\
      0, & \text{otherwise}
    \end{cases},
\end{equation}
with overlap threshold $r_o$. The reverse loss $\mathcal{L}_o^\mathbf{Q}$ is computed in the same way.
The contributions from positive and negative examples are balanced with weights inversely proportional to their relative frequencies.

\parahead{Matchability loss}
Supervising the matchability scores is more difficult, as it is not clear in advance which are the right points to take into account during correspondence search.
We follow a simple intuition: good keypoints are those that can be matched successfully at a given point during training, with the current feature descriptors.
Hence, we cast the prediction as binary classification and generate the ground truth labels on the fly.
Again, we sum the two symmetric losses, $\mathcal{L}_m = \frac{1}{2} (\mathcal{L}^\mathbf{P}_m + \mathcal{L}^\mathbf{Q}_m)$, with
\begin{equation}
\mathcal{L}^\mathbf{P}_m\!= \frac{1}{|\mathbf{P}|} \sum_{i=1}^{|\mathbf{P}|} \overbar{m}_{\mathbf{p}_i}\log(m_{\mathbf{p}_i})+(1-\overbar{m}_{\mathbf{p}_i})\log(1-m_{\mathbf{p}_i}),
\end{equation}
where ground truth labels $\overbar{m}_{\mathbf{p}_i}$ are computed on the fly via nearest neighbour search $\mathsf{NN}_\mathbf{F}(\cdot,\cdot)$ in feature space:
\begin{equation}
    \overbar{m}_{\mathbf{p}_i}\!=     
    \begin{cases}
      1, & ||\overbar{\mathbf{T}}_\mathbf{P}^\mathbf{Q}(\mathbf{p}_i)\!-\!\mathsf{NN}_\mathbf{F}(\mathbf{p}_i,\mathbf{Q}) ||_2\!<\!r_m \\
      0, & \text{otherwise}.
    \end{cases}
\end{equation} 

\begin{figure}[t]
    \centering
    \includegraphics[width=\columnwidth]{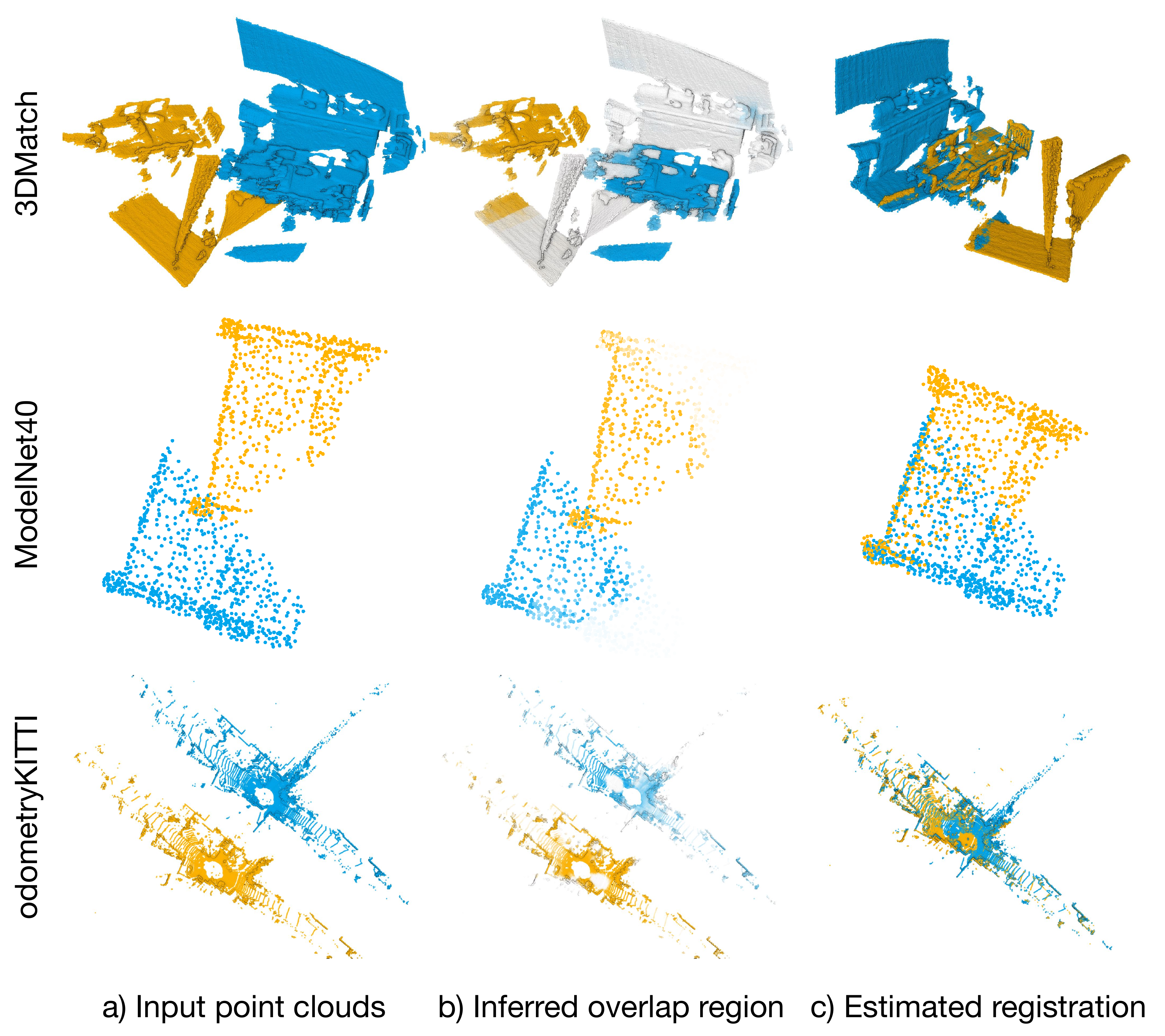}\vspace{-2mm}
    \caption{Example results of \acro\ that succeeds in attending to the overlap region to enable robust registration.}
    \label{fig:3DMatch_qualitative}
    \vspace{-\baselineskip}
\end{figure}

\parahead{Implementation and training}
\acro\ is implemented in pytorch and can be trained on a single RTX 3090 GPU. At the start of the training we supervise \acro\ only with the circle and overlap losses, the matchability loss is added only after few epochs, when the point-wise features are already meaningful (i.e., $>$30\% of interest points can be matched correctly).
The three loss terms are weighted equally. For more details, please refer to Appendix.
\section{Experiments}
\label{sec:experiments}
We evaluate \acro\ and justify our design choices on real point clouds, using \emph{3DMatch}~\cite{zeng20163dmatch} and \emph{3DLoMatch} (\cref{sec:3DMatch}). Additionally, we compare \acro\ to direct registration methods on the synthetic, object-centric \emph{ModelNet40}~\cite{wu2015ModelNet} (\cref{sec:model_net}) and evaluate it on large outdoor scenes using odometryKITTI~\cite{geiger2012kitti} (\cref{sec:kitti}). More details about the datasets and evaluation metrics are available in the Appndix. Qualitative results are shown in Fig.~\ref{fig:3DMatch_qualitative}.

\begin{figure}[t]
    \centering
    \includegraphics[width=\columnwidth]{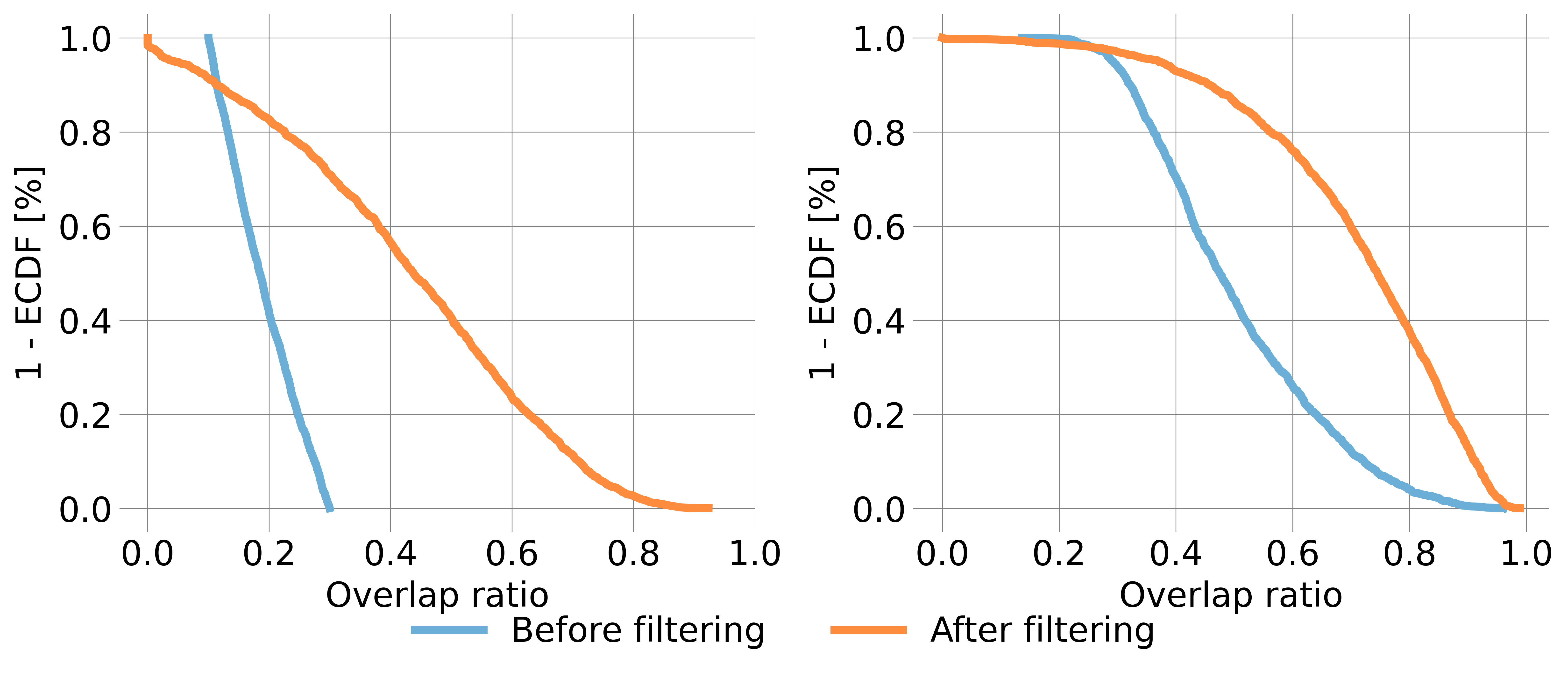}\vspace{-2mm}

    \caption{Distribution of the relative overlap ratio before and after filtering the points with the inferred overlap scores, \textit{3DLoMatch} (left) and \textit{3DMatch} (right).}
    \label{fig:imp_overlap}
    \vspace{-\baselineskip}
\end{figure}
\subsection{3DMatch}
\label{sec:3DMatch}
\parahead{Dataset}
~\cite{zeng20163dmatch} is a collection of 62 scenes, from which we use 46 scenes for training, 8 scenes for validation and 8 for testing. Official \emph{3DMatch} dataset considers only scan pairs with \textgreater30\% overlap. Here, we add its counterpart in which we consider only scan pairs with overlaps between 10 and 30\% and call this collection \textit{3DLoMatch}\footnote{Due to a bug in the official implementation of the overlap computation for \emph{3DMatch}, a few (\textless7\%) scan pairs are included in both datasets.}. 

\parahead{Metrics}
Our main metric, corresponding to the actual aim of point cloud registration, is \emph{Registration Recall~(RR)}, i.e., the fraction of scan pairs for which the correct transformation parameters are found with RANSAC.
Following the literature~\cite{zeng20163dmatch, gojcic2018learned, Choy2019FCGF}, we also report 
\emph{Feature Match Recall~(FMR)}, defined as the fraction of pairs 
that have \textgreater5\% "inlier" matches with \textless10~cm residual under the ground truth transformation (without checking if the transformation can be recovered from those matches), and \emph{Inlier Ratio~(IR)}, the fraction of correct correspondences among the putative matches. Additionally, we use empirical cumulative distribution functions (ECDF) to evaluate the relative overlap ratio. At a specific overlap value, the $(1\!-\!\text{ECDF})$ curve shows the fraction of fragment pairs that have relative overlap greater or equal to that value.

\parahead{Relative overlap ratio}
We first evaluate if \acro\ achieves its goal to focus on the overlap. We discard points with a predicted overlap score
$\mathbf{o}_i\!<\!0.5$, compute the overlap ratio, and compare it to the one of the original scans.
Fig.~\ref{fig:imp_overlap} shows that more than half ($71\%$) of the low-overlap pairs are pushed over the 30\% threshold that prior works considered the lower limit for registration. On average, discarding points with low overlap scores almost doubles the overlap in \emph{3DLoMatch} ($133\%$ increase).
Notably, it also increases the overlap in standard \emph{3DMatch} by, on average, \textgreater50\%.

\begin{figure}[t]
    \centering
    \includegraphics[width=\columnwidth]{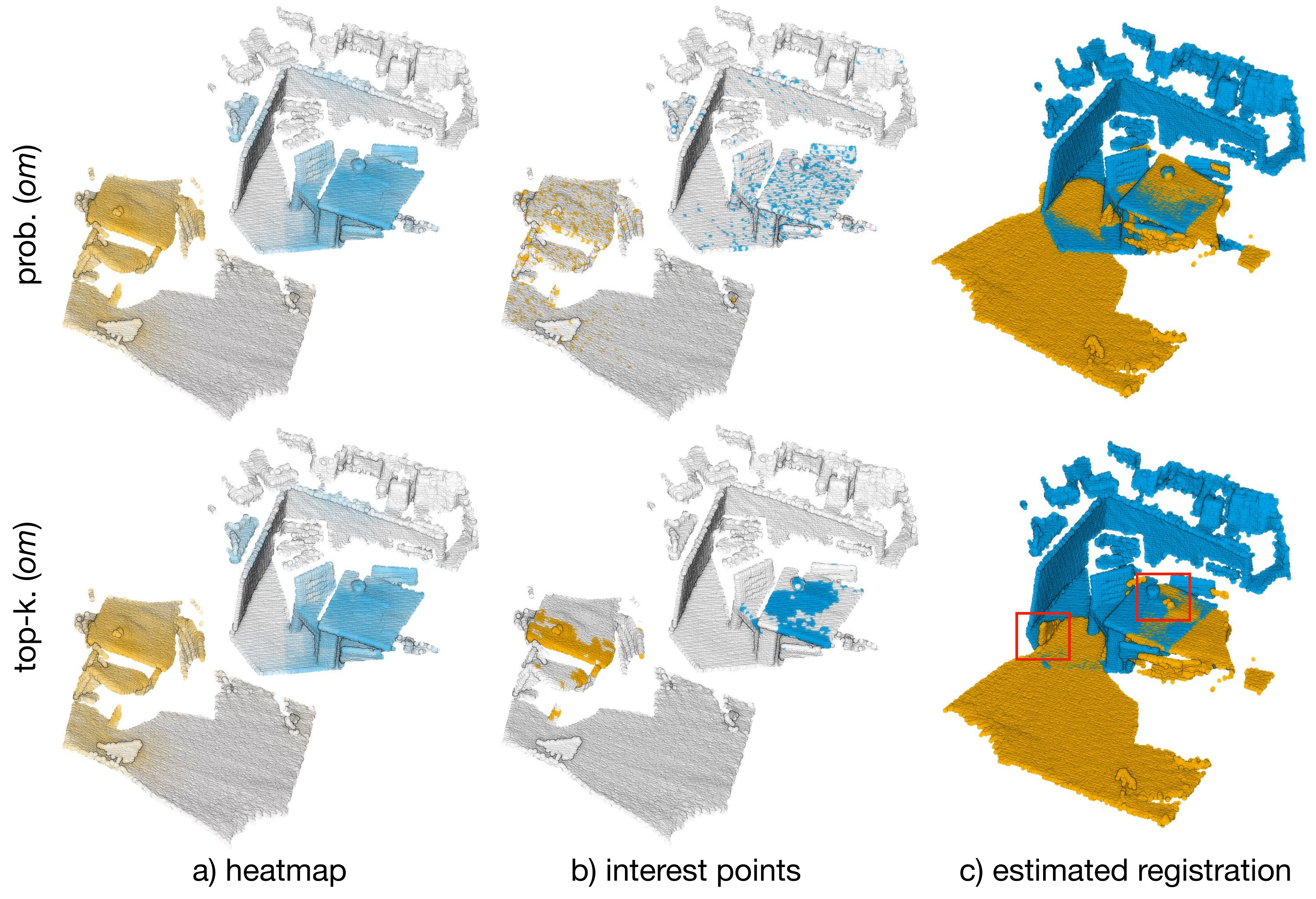}\vspace{-2mm}
    \caption{\emph{Top-k ($om$)} sampling yields clustered interest points, whereas the points obtained with \emph{prob. ($om$)} sampling are more scattered and thus enable a more robust estimation of the transformation parameters.}
    \label{fig:sampling}
    \vspace{-0.5\baselineskip}
\end{figure}

\begin{table}[t]
    \setlength{\tabcolsep}{6pt}
    \renewcommand{\arraystretch}{1.2}
	\centering
	\resizebox{\columnwidth}{!}{
    \begin{tabular}{lccccc|ccccc}
			\toprule
			& \multicolumn{5}{c|}{\textit{3DMatch}} & \multicolumn{5}{c}{\textit{3DLoMatch}} \\

			\# Samples (k) & 5000 & 2500 & 1000 & 500 & 250 & 5000 & 2500 & 1000 & 500 & 250 \\
			\midrule

			& \multicolumn{10}{c}{\textit{Inlier ratio (\%)}} \\
			\midrule
			\emph{rand} & 51.6 & 49.5 & 44.5 & 38.9 & 32.1 & 20.4 & 19.2 & 16.8 & 14.3 & 11.5 \\
 			\emph{top-k ($om$)} & \textbf{68.4} & \textbf{73.8} & \textbf{77.6} & \textbf{78.6} & \textbf{78.7} & \textbf{33.7} & \textbf{39.9} & \textbf{44.9} & \textbf{47.0} & \textbf{47.7} \\
 			\emph{prob. ($om$)} & \underline{58.0} & \underline{58.4} & \underline{57.1} & \underline{54.1} & \underline{49.3} & \underline{26.7} & \underline{28.1} & \underline{28.3} & \underline{27.5} & \underline{25.8} \\

			\midrule
			& \multicolumn{10}{c}{\textit{Registration Recall (\%)}} \\
			\midrule
			\emph{rand} & 86.0 & 84.8 & \underline{84.7} & \underline{81.7} & \underline{75.3} & 43.3 & 45.3 & 40.4 & 35.9 & 28.0 \\
 			\emph{top-k ($om$)} & \underline{88.9} & \underline{87.4} & 82.0 & 75.6 & 64.0 & \underline{58.5} & \underline{57.8} & \underline{53.1} & \underline{44.9} & \underline{35.9} \\
 			\emph{prob. ($om$)} & \textbf{89.0} & \textbf{89.9} & \textbf{90.6} & \textbf{88.5} & \textbf{86.6} & \textbf{59.8} & \textbf{61.2} & \textbf{62.4} & \textbf{60.8} & \textbf{58.1}  \\
			\bottomrule
			
	\end{tabular}
	}
	\vspace{-0.5em}
	\caption{%
	Performance of \acro\ with different interest point sampling strategies; $om$ denotes the product of overlap score and matchability score.
	}
	\label{tab:3DMatch_sampling}
    \vspace{-\baselineskip}
\end{table}
\begin{figure*}[t!]
    \centering
    \includegraphics[width=1\textwidth]{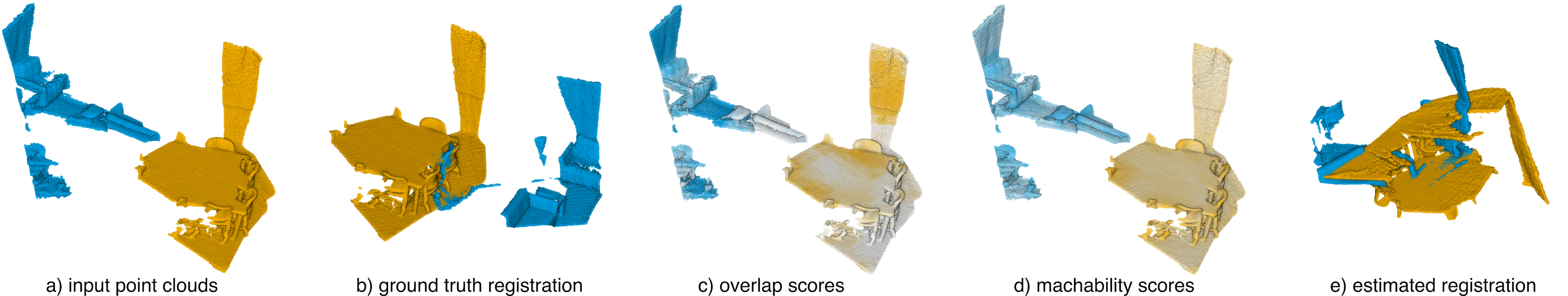}\vspace{-2mm}
    \caption{An extreme case where the overlap is insufficient for registration even with the proposed attention mechanism.}
    \label{fig:failure_image}
    \vspace{-\baselineskip}
\end{figure*}
\parahead{Interest point sampling}
\acro\ significantly increases the effective overlap, but does that improve registration performance?
To test this we use the product of the overlap scores $\mathbf{o}$ and matchability scores $\mathbf{m}$ to bias interest point sampling.
We compare two variants: \emph{top-k (om)}, where we pick the top-$k$ points according to the multiplied scores;
and \emph{prob.\ (om)}, where we instead sample points with probability proportional to the multiplied scores. 

For a more comprehensive assessment we follow~\cite{bai2020d3feat} and report performance with different numbers of sampled interest points. Tab.~\ref{tab:3DMatch_sampling} shows that any of the informed sampling strategies greatly increases the \emph{inlier ratio}, and as a consequence also the \emph{registration recall}.
The gains are larger when fewer points are sampled. In the low-overlap regime the inlier ratios more than triple for up to 1000 points.
We observe that, as expected, high inlier ratio does not necessarily imply high registration recall: our scores are apparently well calibrated, so that \emph{top-k (om)} indeed finds most inliers, but these are often clustered and too close to each other to reliably estimate the transformation parameters (Fig.~\ref{fig:sampling}).
We thus use the more robust \emph{prob. (om)} sampling, which yields the best \emph{registration recall}. It may be possible to achieve even higher registration recall by combining \emph{top-k (om)} sampling with non-maxima suppression. We leave this for future work.

\begin{table}[t]
    \setlength{\tabcolsep}{4pt}
    \renewcommand{\arraystretch}{1.2}
	\centering
	\resizebox{\columnwidth}{!}{
    \begin{tabular}{lccccc|ccccc}
			\toprule
			& \multicolumn{5}{c|}{\textit{3DMatch}} & \multicolumn{5}{c}{\textit{3DLoMatch}} \\
			\# Samples & 5000 & 2500 & 1000 & 500 & 250 & 5000 & 2500 & 1000 & 500 & 250 \\
            \midrule
			& \multicolumn{10}{c}{\textit{Registration Recall (\%)}} \\
		    \midrule
 			3DSN~\cite{gojcic20193DSmoothNet} & 78.4 & 76.2 & 71.4 & 67.6 & 50.8 & 33.0 & 29.0 & 23.3 & 17.0 & 11.0  \\
 			FCGF~\cite{Choy2019FCGF} & \underline{85.1} & \underline{84.7} & 83.3 & 81.6 & 71.4 & \underline{40.1} & 41.7 & 38.2 & 35.4 & 26.8 \\
 			D3Feat~\cite{bai2020d3feat} & 81.6 & 84.5 & \underline{83.4} & \underline{82.4} & \underline{77.9} & 37.2 & \underline{42.7} & \underline{46.9} & \underline{43.8} & \underline{39.1} \\
 			\acro\ & \textbf{89.0} & \textbf{89.9} & \textbf{90.6} & \textbf{88.5} & \textbf{86.6} & \textbf{59.8} & \textbf{61.2} & \textbf{62.4} & \textbf{60.8} & \textbf{58.1} \\
			\bottomrule
			
	\end{tabular}
	}
	\caption{Results on the \emph{3DMatch} and \emph{3DLoMatch} datasets.}
	\label{tab:big_predator}
    \vspace{-\baselineskip}
\end{table}

\parahead{Comparison to feature-based methods}
We compare \acro\ to recent feature-based registration methods: 3DSN~\cite{gojcic2018learned}, FCGF~\cite{Choy2019FCGF} and D3Feat~\cite{bai2020d3feat}, see Tab.~\ref{tab:big_predator}. 
Even though \acro\ can not solve all the cases (\cf Fig.~\ref{fig:failure_image}), it greatly outperforms existing methods on the low-overlap \emph{3DLoMatch} dataset, improving registration recall by 15.5-19.7 percent points (pp) over the closest competitor---variously FCGF or 3DFeat. 
Moreover, it also consistently reaches the highest registration recall on standard \emph{3DMatch}, showing that its attention to the overlap pays off even for scans with moderately large overlap.
In line with our motivation, what matters is not so much the choice of descriptors, but finding interest points that lie in the overlap region -- especially if that region is small. 
%Additionally, we show that a larger network (see big\acro\ in Tab.~\ref{tab:big_predator}, with 2$\times$ bigger network width) can further boost the performance.

\parahead{Comparison to direct registration methods}
We also tried to compare \acro\ to recent methods for direct registration of partial point clouds.
Unfortunately, for both PRNet~\cite{wang2019prnet} and RPM-Net~\cite{yew2020rpm}, training on \emph{3DMatch} failed to converge to reasonable results, as already observed in~\cite{choy2020deep}.
It appears that their feature extraction is specifically tuned to synthetic, object-centric point clouds.
Thus, in a further attempt we replaced the feature extractor of RPM-Net with FCGF.
This brought the registration recall on \emph{3DMatch} to 54.9\%, still far from the
85.1\%  that FCGF features achieve with RANSAC.
We conclude that direct pairwise registration is at this point only suitable for geometrically simple objects in controlled settings like \emph{ModelNet40}.

\begin{table}[t!]
    \setlength{\tabcolsep}{10pt}
    \renewcommand{\arraystretch}{1.2}
	\centering
	\resizebox{\columnwidth}{!}{
    \begin{tabular}{lll|cccccc}
			\toprule
			\multicolumn{3}{c}{overlap attention} & \multicolumn{3}{c}{\textit{3DMatch}} & \multicolumn{3}{c}{\textit{3DLoMatch}} \\
            \cline{4-9}
            \multicolumn{1}{c}{\textit{ov.}} & \multicolumn{1}{c}{\textit{$\times$ov.}} & \multicolumn{1}{c}{\textit{cond.}} & FMR & IR & RR & FMR & IR & RR \\
            \hline
            \multicolumn{1}{c}{}& \multicolumn{1}{c}{} & \multicolumn{1}{c}{} & \underline{96.4} & 39.6 & 82.6 & 72.2 & 14.5 & 38.9\\
            \multicolumn{1}{c}{\ding{51}} & \multicolumn{1}{c}{} & \multicolumn{1}{c}{} & 94.6 & 38.3 & 84.1 & 67.1 & 14.3 & 42.8 \\
            \multicolumn{1}{c}{\ding{51}} & \multicolumn{1}{c}{\ding{51}} & \multicolumn{1}{c}{} & \underline{96.4} & 50.8 & 87.7 & \underline{73.8} & 20.9 & 56.5 \\
            \multicolumn{1}{c}{\ding{51}} & \multicolumn{1}{c}{}  & \multicolumn{1}{c}{\ding{51}} &  95.7 & \underline{52.1} & \underline{88.0} & 72.5 & \underline{21.2} & \underline{57.5} \\
            \multicolumn{1}{c}{\ding{51}} & \multicolumn{1}{c}{\ding{51}} & \multicolumn{1}{c}{\ding{51}} & \textbf{96.7} & \textbf{58.0} & \textbf{89.0} & \textbf{78.6} & \textbf{26.7}& \textbf{59.8} \\
			\bottomrule
	\end{tabular}
	}
	\caption{Ablation of the network architecture. \textit{ov.} denotes upsampling the overlap scores; \textit{cond.} denotes conditioning the bottleneck features on the respective other point cloud; \textit{$\times$ov.} denotes upsampling the cross overlap scores.}
	\label{tab:3DMatch_ablation_w_baseline}
	\vspace{-\baselineskip}
\end{table}

\parahead{Ablations study}
We ablate our overlap attention module in Tab.~\ref{tab:3DMatch_ablation_w_baseline}.
We first compare \acro\ with a baseline model, in which we completely remove the proposed overlap attention module. That baseline, combined with random sampling, achieves the 2\textsuperscript{nd}-highest FMR on both benchmarks, but only reaches 82.6\%, respectively 38.9\% RR.
%; much worse than the other variants that include (at least) the overlap scores. 
%The experiment again confirms that high FMR or IR does not imply high RR, and thus good registration performance. 
By adding the overlap scores, RR increases by 1.5, respectively 3.9 pp on \emph{3DMatch} and \emph{3DLoMatch}. Additionally upsampling conditioned feature scores or cross overlap scores further improves performance, especially on \emph{3DLoMatch}. All three parts combined lead to the best overall performance. For further ablation studies, see Appendix.

\subsection{ModelNet40}
\label{sec:model_net}
\parahead{Dataset}
~\cite{wu2015ModelNet} contains 12,311 CAD models of man-made objects from 40 different categories. We follow~\cite{yew2020rpm} to use 5,112 samples for training, 1,202 samples for validation, and 1,266 samples for testing. Partial scans are generated following~\cite{yew2020rpm}. In addition to \emph{ModelNet} which has 73.5\% pairwise overlap on average, we generate \emph{ModelLoNet} with lower (53.6\%) average overlap. For more details see Appendix. 

\parahead{Metrics}
We follow \cite{yew2020rpm} and measure the performance using the \emph{Relative Rotation Error (RRE)} (geodesic distance between estimated and GT rotation matrices), the \emph{Relative Translation Error (RTE)} (Euclidean distance between the estimated and GT translations), and the \emph{Chamfer distance (CD)} between the two registered scans.

\parahead{Relative overlap ratio} 
\begin{figure}[t]
    \centering
    \includegraphics[width=\columnwidth]{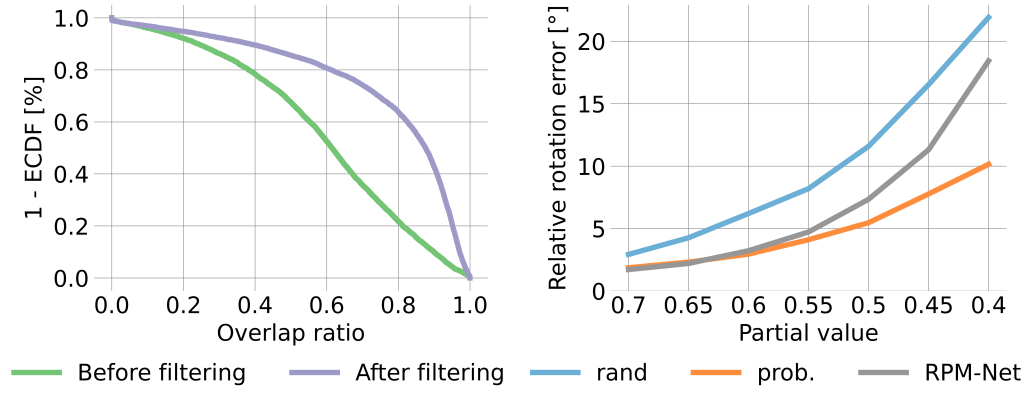}\vspace{-2mm}

    \caption{Improved relative overlap ratio after filtering the points with the inferred overlap scores on 8862 \textit{ModelNet} partial scans(\textit{left}). Owing to the improved overlap ratio, \acro\ is robust to the changes of partial value $p_v$, while the performance of RPM-Net drops rapidly (\textit{right}).  \textit{rand} and \textit{prob.} denote the random and \emph{prob. (om)} biased sampling of 450 interest points, respectively.}%
    \label{fig:modelnet}
    \vspace{-0.5\baselineskip}
\end{figure}
We again evaluate if \acro\ focuses on the overlap region. We extract 8,862 test pairs by varying the completeness of the input point clouds from 70 to 40\%. Fig.~\ref{fig:modelnet} shows that \acro\ substantially increases the relative overlap and reduces the number of pairs with overlap \textless70\% by more than 40 pp.

\begin{table}[t]
    \setlength{\tabcolsep}{4pt}
    \renewcommand{\arraystretch}{1.2}
	\centering
	\resizebox{\columnwidth}{!}{
    \begin{tabular}{lcccccc}
			\hline
			& \multicolumn{3}{c}{\textit{ModelNet}} & \multicolumn{3}{c}{\textit{ModelLoNet}} \\
			\cline{2-7}
			Methods & RRE & RTE & $CD$ & RRE & RTE & $CD$ \\
			\hline
 			DCP-v2~\cite{wang2019dcp} & 11.975 & 0.171 & 0.0117 & 16.501 & 0.300 & 0.0268 \\
 			RPM-Net~\cite{yew2020rpm} & \textbf{1.712} & \textbf{0.018} & \textbf{0.00085} & \underline{7.342} & \textbf{0.124} & \textbf{0.0050}\\
 			\acro\ (\emph{rand}) & 2.407 & 0.028 & 0.00120 & 10.985 & 0.175 & 0.0097 \\ 
 			\acro\ (\emph{prob. ($om$)})  & \underline{1.739} & \underline{0.019} & \underline{0.00089} & \textbf{5.235} & \underline{0.132} & \underline{0.0083} \\
			\hline
	\end{tabular}
	}
	\caption{Evaluation results on \emph{ModelNet} and \emph{ModelLoNet}. 450 points are sampled for RANSAC with \textit{rand} / \textit{prob.}.}
	\label{tab:modelnet}
	\vspace{-0.8\baselineskip}
\end{table}

\parahead{Comparison to direct registration methods}
To be able to compare \acro\ to RPM-Net~\cite{yew2020rpm} and DCP~\cite{wang2019dcp}, we resort to the synthetic, object-centric dataset they were designed for. We failed to train PRNet~\cite{wang2019prnet} due to random crashes of the original code (also observed in~\cite{choy2020deep}).

Remarkably, \acro\ can compete with methods specifically tuned for \emph{ModelNet}, and in the low-overlap regime outperforms them in terms of \emph{RRE}, see Tab.~\ref{tab:modelnet}.
Moreover, we observe a large boost by sampling points with overlap attention (\emph{prob.\ (om)}) rather than randomly (\emph{rand}).
Fig.~\ref{fig:modelnet} (right) further underlines the importance of sampling in the overlap: \acro\ is a lot more robust in the low overlap regime ($\approx$8$^\circ$ lower RRE at completeness 0.4). 

\subsection{odometryKITTI}
\label{sec:kitti}
\parahead{Dataset}
~\cite{geiger2012kitti} contains 11 sequences of LiDAR-scanned outdoor driving scenarios. We follow ~\cite{Choy2019FCGF} and use sequences 0-5 for training, 6-7 for validation, and 8-10 for testing. In line with~\cite{Choy2019FCGF, bai2020d3feat} we further refine the provided ground truth poses using ICP~\cite{besl1992method} and only use point cloud pairs that are at most $10$~m away from each other for evaluation.

\parahead{Comparision to the SoTAs} 
\begin{table}[t]
    \setlength{\tabcolsep}{11pt}
    \renewcommand{\arraystretch}{1.05}
	\centering
	\resizebox{\columnwidth}{!}{
    \begin{tabular}{lccccc}
			\hline
			Method & \textit{RTE [cm]}~$\downarrow$ & \textit{RRE [$^\circ$]}~$\downarrow$ & RR~$\uparrow$  \\
			\hline
 			3DFeat-Net~\cite{yew20183dfeat} & 25.9 &  0.57 & 96.0  \\
 			FCGF~\cite{Choy2019FCGF} & 9.5  & 0.30  & 96.6\\
 			D3Feat*~\cite{bai2020d3feat} & \underline{7.2} %
 			& \underline{0.30} %
 			& \textbf{99.8} \\
 			\acro\ (\textit{rand}) & 8.8 & 0.34 & \textbf{99.8}  \\ 
 			\acro\ (\textit{prob. (om)}) &  \textbf{6.8} & \textbf{0.27} & \textbf{99.8} \\
			\hline
	\end{tabular}
	}
	\vspace{0.3mm}
	\caption{Evaluation of \acro\ on \emph{odometryKITTI}, following the evaluation protocol employed by D3Feat~\cite{bai2020d3feat}.}
	\vspace{-2mm}
	\label{tab:kitti}
    \vspace{-0.5\baselineskip}
\end{table}
We compare \acro\ to 3DFeat-Net~\cite{yew20183dfeat}, FCGF~\cite{Choy2019FCGF} and D3Feat*~\cite{bai2020d3feat}\footnote{We find that the released D3Feat code fails to reproduce the results in the paper, possible due to hyper-parameter changes.}
%on \emph{odometryKITTI}. 
As shown in Tab.~\ref{tab:kitti}, \acro\ performs on-par with the SoTA. The results also corroborate the impact of our overlap attention %(\textit{prob.(om)}), 
which again outperforms the random sampling baseline.
%(\textit{rand}). 

\parahead{Computational complexity}
With $O(n^2)$ complexity the cross-attention module represents the memory bottleneck of \acro. Furthermore, $n$ cannot be selected freely but results from the interplay of (i) the resolution of the initial voxel grid, (ii) the network architecture (number of strided convolution layers), and (iii) the spatial extent of the scene. Nevertheless, by executing the cross-attention at the \emph{superpoint} level, with greatly reduced $n$, we are able to apply \acro\ to large outdoor scans like \emph{odometryKITTI} on a single GPU. For even larger scenes, a simple engineering trick could be to split them into parts, as often done for semantic segmentation.

\vspace{-2mm}
\section{Conclusion}
We have %
introduced \acro, a deep model designed for pairwise registration of low-overlap point clouds. The core of the model is an overlap attention module that enables early information exchange between the point clouds' latent encodings, in order to infer which of their points are likely to lie in their overlap region.

There are a number of directions in which \acro\ could be extended. At present it is tightly coupled to fully convolutional point cloud encoders, and relies on having a reasonable number of \emph{superpoints} in the bottleneck. This could be a limitation in scenarios where the point density is very uneven. It would also be interesting to explore how our overlap-attention module can be integrated into direct point cloud registration methods and other neural architectures that have to handle two inputs with low overlap, e.g. in image matching~\cite{sarlin2020superglue}. Finally, registration in the low-overlap regime is challenging and \acro\ cannot solve all the cases. A user study could provide a better understanding of how \acro\ compares to human operators.

\noindent\textbf{Acknowledgements.}
{This work was sponsored by the NVIDIA GPU grant.}

{\small
\bibliographystyle{ieee_fullname.bst}
\bibliography{main.bib}
}

\newpage
\setcounter{section}{0}
\renewcommand\thesection{\Alph{section}}
\section{Appendix}
{\let\thefootnote\relax\footnote{{$^*$First two authors contributed equally to this work.}}}
In this supplementary material, we first provide rigorous definitions of evaluation metrics (Sec.~\ref{sec:evaluation_metrics_supp}), then describe the data pre-processing step (Sec.~\ref{sec:datasets_supp}), network architectures (Sec.~\ref{sec:network_arch_supp}) and training on individual datasets (Sec.~\ref{sec:training_supp}) in more detail. We further provide additional results (Sec.~\ref{sec:additional_results_supp}), ablation studies (Sec.~\ref{sec:addtional_ablation_supp}) as well as a runtime analysis (Sec.~\ref{sec:timing}). Finally, we show more visualisations on \emph{3DLoMatch} and \emph{ModelLoNet} benchmarks (Sec.~\ref{sec:qualitative_supp}).

\subsection{Evaluation metrics}
\label{sec:evaluation_metrics_supp}
The evaluation metrics, which we use to assess model performance in Sec.~4 of the main paper and Sec.~\ref{sec:additional_results_supp} of this supplementary material, are formally defined as follows:

\textbf{Inlier ratio} looks at the set of putative correspondences $(\mathbf{p}, \mathbf{q}) \in \mathcal{K}_{i j}$ found by reciprocal matching%
in feature space, and measures what fraction of them is "correct", in the sense that they lie within a threshold $\tau_1\!=\!10\,$cm after registering the two scans with the ground truth transformation $\overbar{T}_\mathbf{P}^ \mathbf{Q}$:
\begin{equation}
\mathrm{IR} = \frac{1}{\left|\mathcal{K}_{ij}\right|} \sum_{\left(\mathbf{p}, \mathbf{q}\right) \in \mathcal{K}_{ij}} \big[ ||\overbar{\mathbf{T}}_\mathbf{P}^ \mathbf{Q}(\mathbf{p})-\mathbf{q}||_2<\tau_{1} \big]  \;,
\end{equation}
with $[\cdot]$ the Iverson bracket.

\textbf{Feature Match recall} (FMR)~\cite{deng2018ppfnet} measures the fraction of point cloud pairs for which, based on the number of inlier correspondences, it is \emph{likely} that accurate transformation parameters can be recovered with a robust estimator such as RANSAC.
Note that FMR only checks whether the inlier ratio is above a threshold $\tau_2=0.05$. It does not test if the transformation can actually be determined from those correspondences, which in practice is not always the case, since their geometric configuration may be (nearly) degenerate, e.g., they might lie very close together or along a straight edge.
A single pair of point clouds counts as suitable for registration if%
\begin{equation}
    IR > \tau_2
\end{equation}

\textbf{Registration recall}~\cite{choi2015robust} is the most reliable metric, as it measures end-to-end performance on the actual task of point cloud registration. Specifically, it looks at the set of ground truth correspondences $\mathcal{H}_{i j}^{*}$ after applying the estimated transformation $T_\mathbf{P}^ \mathbf{Q}$, computes their root mean square error,
\begin{equation}
    \mathrm{RMSE} = \sqrt{\frac{1}{\left|\mathcal{H}_{i j}^{*}\right|} \sum_{\left(\mathbf{p}, \mathbf{q}\right) \in \mathcal{H}_{i j}^{*}}||\mathbf{T}_\mathbf{P}^ \mathbf{Q}(\mathbf{p}) -\mathbf{q}||_2^2}\;,
\end{equation}
and checks for what fraction of all point pairs $\mathrm{RMSE}\!<\!0.2$.
In keeping with the original evaluation script of \emph{3DMatch}, immediately adjacent point clouds are excluded, since they have very high overlap by construction.

\textbf{Chamfer distance} measures the quality of registration on synthetic data. We follow ~\cite{yew2020rpm} and use the \emph{modified} Chamfer distance metric:
\begin{equation}
\begin{aligned}
\tilde{C D}(\mathbf{P}, \mathbf{Q}) = & \frac{1}{|\mathbf{P}|} \sum\limits_{\mathbf{p} \in \mathbf{P}} \min\limits_{\mathbf{q} \in \mathbf{Q}_{\text{raw}}}\|\mathbf{T}_\mathbf{P}^ \mathbf{Q}(\mathbf{p})-\mathbf{q}\|_{2}^{2} + \\
& \frac{1}{|\mathbf{Q}|} \sum\limits_{\mathbf{q} \in \mathbf{Q}} \min\limits_{\mathbf{p} \in \mathbf{P}_{\text{raw}}}\|\mathbf{q}- \mathbf{T}_\mathbf{P}^ \mathbf{Q}(\mathbf{p})\|_{2}^{2}
\end{aligned}
\end{equation}
where $\mathbf{P}_{\text{raw}}\in \mathbb{R}^{2048\times3}$ and $\mathbf{Q}_{\text{raw}}\in \mathbb{R}^{2048\times3}$ are \emph{raw} source and target point clouds, $\mathbf{P} \in \mathbb{R}^{717\times3}$ and $\mathbf{Q} \in \mathbb{R}^{717\times3}$ are \emph{input} source and target point clouds. 

\textbf{Relative translation and rotation errors} (RTE/RRE) measures the deviations from the ground truth pose as: 
\begin{equation}
\begin{aligned}
\text{RTE} &= \|\mathbf{t} - \overbar{\mathbf{t}}\|_2\\
\text{RRE} &=\arccos\big(\frac{\mathrm{trace}{(\mathbf{R}^{T}\overbar{\mathbf{R}})-1}}{2}\big)
\end{aligned}
\end{equation}
where $\mathbf{R}$ and $\mathbf{t}$ denote the estimated rotation matrix and translation vector, respectively.

\textbf{Empirical Cumulative Distribution Function} (ECDF) measures the distribution of a set of values:
\begin{equation}
\begin{aligned}
\text{ECDF} (x) = \frac{\big|\{o_i < x\}\big|}{\big|O\big|}
\end{aligned}
\end{equation}
where $O$ is a set of values(ovelap ratios in our case) and $x \in [\min\{O\}, \max\{O\}]$.

\subsection{Dataset preprocessing}
\label{sec:datasets_supp}
\parahead{3DMatch}
\cite{zeng20163dmatch} is a collection of 62 scenes, combining earlier data from Analysis-by-Synthesis~\cite{valentin2016learning}, 7Scenes~\cite{shotton2013scene}, SUN3D~\cite{xiao2013sun3d}, RGB-D Scenes v.2~\cite{lai2014unsupervised}, and Halber~\etal~\cite{Halber2016StructuredGR}.  The official benchmark splits the data into 54 scenes for training and 8 for testing. Individual scenes are not only captured in different indoor spaces (e.g., bedrooms, offices, living rooms, restrooms) but also with different depth sensors (e.g., Microsoft Kinect, Structure Sensor, Asus Xtion Pro Live, and Intel RealSense). \emph{3DMatch} provides great diversity and allows our model to generalize across different indoor spaces. Individual scenes of \emph{3DMatch} are split into point cloud fragments, which are generated by fusing 50 consecutive depth frames using TSDF volumetric fusion~\cite{curless1996volumetric}. As a preprocessing step, we apply voxel-grid downsampling to all point clouds, and if multiple points fall into the same voxel, we randomly pick one.

\parahead{ModelNet40}
For each CAD model of \emph{ModelNet40}, 2048 points are first generated by uniform sampling and scaled to fit into a unit sphere. Then we follow~\cite{yew2020rpm} to produce partial scans: for source partial point cloud, we uniformly sample a plane through the origin that splits the unit sphere into two half-spaces, shift that plane along its normal until $\lfloor 2048\cdot p_v \rfloor$ points are on one side, and discard the points on the other side; the target point cloud is generated in the same manner; then the two resulting, partial point clouds are randomly rotated, translated and jittered with Gaussian noise. For the rotation, we sample a random axis and a random angle \textless45$^\circ$. The translation is sampled in the range $[-0.5,0.5]$. Gaussian noise is applied per coordinate with $\sigma\!=\!0.05$. Finally, 717 points are randomly sampled from the $\lfloor 2048\cdot p_v \rfloor$ points.

\parahead{odometryKITTI}
The dataset was captured using a Velodyne HDL-64 3D laser scanner by driving around the mid-size city of Karlsruhe, in rural areas and on highways. The ground truth poses are provided by GPS/IMU system. We follow ~\cite{bai2020d3feat} to use ICP to reduce the noise in the ground truth poses.

\subsection{Implementation and training}
\begin{table}[t]
    \setlength{\tabcolsep}{6pt}
    \renewcommand{\arraystretch}{1.2}
	\centering
	\resizebox{\columnwidth}{!}{
    \begin{tabular}{cccccccc}
			\toprule
             & $n_p$ & $\gamma$ & $V$ & $r_p$ & $r_s$ & $r_o$ & $r_m$\\
            \hline
            \multicolumn{1}{c}{\emph{3DMatch}} & 256 & 24 & 0.025 & 0.0375 & 0.1 & 0.0375 & 0.05\\
            \multicolumn{1}{c}{\emph{ModelNet}} & 384 & 64 & 0.06 & 0.018 & 0.06 & 0.04 & 0.04 \\
            \multicolumn{1}{c}{\emph{odometryKITTI}} & 512 & 48 & 0.3 & 0.21 & 0.75 & 0.45 & 0.3 \\
			\bottomrule
			
	\end{tabular}
	}
	\caption{Hyper-parameters configurations for different datasets.}
	\label{tab:hyperparameters}
	\vspace{-\baselineskip}
\end{table}
\label{sec:training_supp}
For 3DMatch/Modelnet/KITTI, we train \acro\ using Stochastic Gradient Descent for $30$/ $200$/ $150$ epochs, with initial learning rate $0.005$/ $0.01$/ $0.05$, momentum $0.98$, and weight decay $10^{-6}$. The learning rate is exponentially decayed by 0.05 after each epoch. Due to memory constraints we use batch size $1$ in all experiments. The dataset-dependent hyper-parameters which include number of negative pairs in circle loss $n_p$, temperature factor $\gamma$, voxel size $V$, search radius for positive pair $r_p$, safe radius $r_s$, overlap and matchability radius $r_o$ and $r_m$ are given in Tab.~\ref{tab:hyperparameters}. On odometryKITTI dataset, we take the curriculum learning~\cite{bengio2009curriculum} strategy to gradually learn sharper local descriptors by adjusting $n_p$. For more details please see our code.

\subsection{Network architecture}
\label{sec:network_arch_supp}
\begin{figure*}[t]
    \centering
    \includegraphics[width=0.95\textwidth]{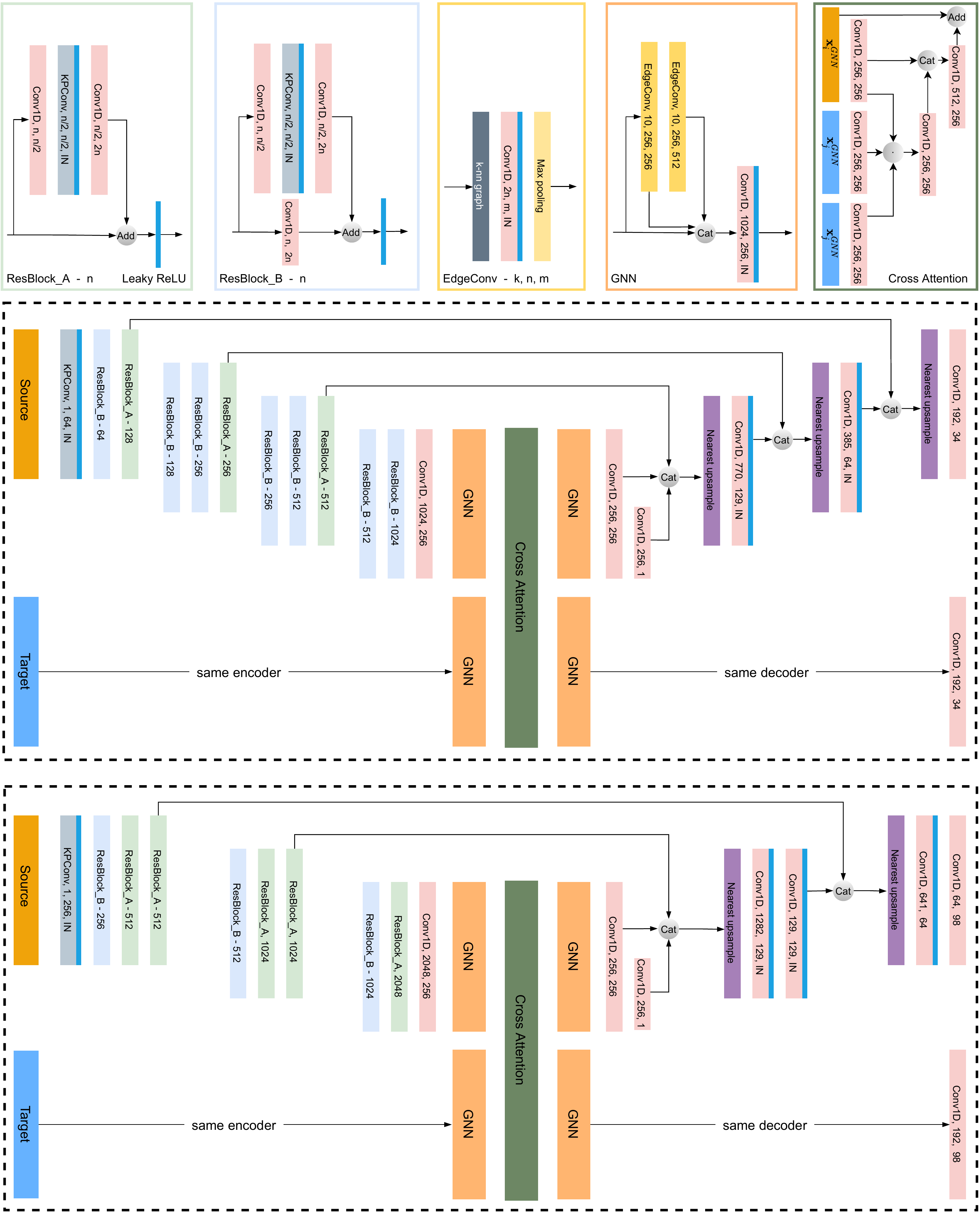}
    \caption{Network architecture of \acro\ for \emph{3DMatch} (\textit{middle}) and \emph{ModelNet} (\textit{bottom}). In the cross attention module, for each (query $\mathbf{s}_i \in \mathbb{R}^{b\times1}$ , key $\mathbf{k}_i \in \mathbb{R}^{b\times1}$, value $\mathbf{v}_i \in \mathbb{R}^{b\times1}$), $\bigodot$ denotes first reshape them into shape $(4, \frac{b}{4})$(4 heads), then compute scores matrix $\mathbf{S}$ from $\mathbf{s}_i$ and $\mathbf{k}_i$, finally get message update from $\mathbf{v}_i$ and reshape back to $(b,1)$.}
    \label{fig:network_arch_detailed}
    \vspace{-\baselineskip}
\end{figure*}

The detailed network architecture of \acro\ is depicted in Fig.~\ref{fig:network_arch_detailed}. Our model is built on the KPConv implementation from the D3Feat repository.%
\footnote{\url{https://github.com/XuyangBai/D3Feat.pytorch}} %
We complement each KPConv layer with instance normalisation Leaky ReLU activations. The $l$-th strided convolution is applied to a point cloud dowsampled with voxel size $2^{l}\cdot V$. Upsampling in the decoder is performed by querying the associated feature of the closest point from the previous layer.  

With $\approx$20k points after voxel-grid downsampling, the point clouds in \emph{3DMatch}  are much denser than those of \emph{ModelNet40} with only 717 points. Moreover, they also have larger spatial extent with bounding boxes up to $3\times3\times3$~$\text{m}^3$, while \emph{ModelNet40} point clouds are normalised to fit into a unit sphere.
To account for these large differences, we slightly adapt the encoder and decoder per dataset, but keep the same overlap attention model. Differences in network hyper-parameters are shown in Tab.~\ref{tab:compare_network}. 
\begin{table}[t]
    \setlength{\tabcolsep}{6pt}
    \renewcommand{\arraystretch}{1.2}
	\centering
	\resizebox{\columnwidth}{!}{
    \begin{tabular}{ccccc}
			\toprule
            \multicolumn{1}{c}{} & \multicolumn{1}{c}{\# strided} & \multicolumn{1}{c}{convolution} & \multicolumn{1}{c}{first conv.} & \multicolumn{1}{c}{final}\\
            \multicolumn{1}{c}{} & \multicolumn{1}{c}{convolutions} & \multicolumn{1}{c}{radius} & \multicolumn{1}{c}{feature dim.} & \multicolumn{1}{c}{feature dim.}\\
            \hline
            \multicolumn{1}{c}{\emph{3DMatch}} & 3 & 2.5 & 64 & 32 \\
            \multicolumn{1}{c}{\emph{ModelNet}} & 2 & 2.75 & 256 & 96 \\
            \multicolumn{1}{c}{\emph{odometryKITTI}} & 3 & 4.25 & 128 & 32 \\
			\bottomrule
	\end{tabular}}
	\caption{Different network configurations for \emph{3DMatch}, \emph{ModelNet} and \emph{odometryKITTI} datasets.}
	\label{tab:compare_network}
\end{table}

\begin{table*}[t!]
    \setlength{\tabcolsep}{4pt}
    \renewcommand{\arraystretch}{1.2}
	\centering
	\resizebox{\linewidth}{!}{
    \begin{tabular}{lcccccccccc|cccccccccc}
			\toprule
			& \multicolumn{10}{c|}{\textit{3DMatch}} & \multicolumn{10}{c}{\textit{3DLoMatch}} \\
			& Kitchen & Home 1 & Home 2 & Hotel 1 & Hotel 2 & Hotel 3 & Study & MIT Lab & Avg. & STD	& Kitchen & Home 1 & Home 2 & Hotel 1 & Hotel 2 & Hotel 3 & Study & MIT Lab & Avg. & STD\\
            \midrule
            & \multicolumn{20}{c}{\textit{\# Sample}} \\
            \midrule
            & \textbf{449} & 106 & 159 & 182 & 78 & 26 & \underline{234} & 45 & 160 & 128 & \textbf{524} & \underline{283} & 222 & 210 & 138 & 42 & 237 & 70 & 191 & 154\\
			\midrule
			& \multicolumn{20}{c}{\textit{Registration Recall (\%)}~$\uparrow$} \\
			\midrule
			3DSN~\cite{gojcic20193DSmoothNet} & 90.6 & 90.6 & 65.4 & 89.6 & 82.1 & 80.8 & 68.4 & 60.0 & 78.4 & 11.5 & 51.4 & 25.9 & 44.1 & 41.1 & 30.7 & 36.6 & 14.0 & 20.3 & 33.0 & 11.8 \\
			FCGF~\cite{Choy2019FCGF} & \textbf{98.0} & \underline{94.3} & \underline{68.6} & \textbf{96.7} & \underline{91.0} & \textbf{84.6} & 76.1 & \underline{71.1} & \underline{85.1} & 11.0 & \underline{60.8} & \underline{42.2} & \underline{53.6} & \underline{53.1} & \underline{38.0} & 26.8 & \underline{16.1} & 30.4 & \underline{40.1} & 14.3\\
			D3Feat~\cite{bai2020d3feat} & 96.0 & 86.8 & 67.3 & 90.7 & 88.5 & 80.8 & \underline{78.2} & 64.4 & 81.6 & \underline{10.5} & 49.7 & 37.2 & 47.3 & 47.8 & 36.5 & \underline{31.7} & 15.7 & \underline{31.9} & 37.2 & \textbf{10.6}\\
			Ours & \underline{97.6} & \textbf{97.2} & \textbf{74.8} & \textbf{98.9} & \textbf{96.2} & \textbf{88.5} & \textbf{85.9} & \textbf{73.3} & \textbf{89.0} & \textbf{9.6} & \textbf
			{71.5} & \textbf{58.2} & \textbf{60.8} & \textbf{77.5} & \textbf{64.2} & \textbf{61.0} & \textbf{45.8} & \textbf{39.1} & \textbf{59.8} & \underline{11.7} \\
			\midrule
			& \multicolumn{20}{c}{\textit{Relative Rotation Error (\degree)}~$\downarrow$} \\
			\midrule
			3DSN~\cite{gojcic20193DSmoothNet} & 1.926 & 1.843 & \underline{2.324} & 2.041 & 1.952 & 2.908 & 2.296 & 2.301 & 2.199 & 0.321 & \underline{3.020} & 3.898 & 3.427 & 3.196 & 3.217 & 3.328 & 4.325 & 3.814 & 3.528 & 0.414\\
			FCGF~\cite{Choy2019FCGF} & \textbf{1.767} & \underline{1.849} & \textbf{2.210} & \textbf{1.867} & \underline{1.667} & \underline{2.417} & \textbf{2.024} & \textbf{1.792} & \textbf{1.949} & \underline{0.236} & \textbf{2.904} & \underline{3.229} & \underline{3.277} & \underline{2.768} & \textbf{2.801} & \textbf{2.822} & \underline{3.372} & 4.006 & \underline{3.147} & 0.394\\
			D3Feat~\cite{bai2020d3feat} & 2.016 & 2.029 & 2.425 & \underline{1.990} & 1.967 & \textbf{2.400} & 2.346 & 2.115 & 2.161 & \textbf{0.183} & 3.226 & 3.492 & 3.373 & 3.330 & 3.165 & \underline{2.972} & 3.708 & \underline{3.619} & 3.361 & \textbf{0.227} \\
			Ours & \underline{1.861} & \textbf{1.806} & 2.473 & 2.045 & \textbf{1.600} & 2.458 & \underline{2.067} & \underline{1.926} & \underline{2.029} & 0.286 & 3.079 & \textbf{2.637} & \textbf{3.220} & \textbf{2.694} & \underline{2.907} & 3.390 & \textbf{3.046} & \textbf{3.412} & \textbf{3.048} & \underline{0.273}\\
			\midrule
			& \multicolumn{20}{c}{\textit{Relative Translation Error (m)}~$\downarrow$} \\
			\midrule
			3DSN~\cite{gojcic20193DSmoothNet} & 0.059 & 0.070 & 0.079 & 0.065 & 0.074 & 0.062 & 0.093 & 0.065 & 0.071 & \textbf{0.010} & \underline{0.082} & 0.098 & 0.096 & 0.101 & \textbf{0.080} & 0.089 & 0.158 & \textbf{0.120} & 0.103 & 0.024\\
			FCGF~\cite{Choy2019FCGF} & \underline{0.053} & \underline{0.056} & \underline{0.071} & \textbf{0.062} & \underline{0.061} & \underline{0.055} & \underline{0.082} & 0.090 & \underline{0.066} & 0.013 & 0.084 & \underline{0.097} & \textbf{0.076} & 0.101 & \underline{0.084} & \underline{0.077} & \underline{0.144} & 0.140 & \underline{0.100} & 0.025\\
			D3Feat~\cite{bai2020d3feat} & 0.055 & 0.065 & 0.080 & \underline{0.064} & 0.078 & \textbf{0.049} & 0.083 & \underline{0.064} & 0.067 & 0.011 & 0.088 & 0.101 & 0.086 & \underline{0.099} & 0.092 & \textbf{0.075} & 0.146 & 0.135 & 0.103 & \underline{0.023}\\
			Ours & \textbf{0.048} & \textbf{0.055} & \textbf{0.070} & 0.073 & \textbf{0.060} & 0.065 & \textbf{0.080} & \textbf{0.063} & \textbf{0.064} & \underline{0.010} & \textbf{0.081} & \textbf{0.080} & \underline{0.084} & \textbf{0.099} & 0.096 & 0.077 & \textbf{0.101} & \underline{0.130} & \textbf{0.093} & \textbf{0.016}\\
			\bottomrule
	\end{tabular}
	}
	\caption{Detailed results on the \emph{3DMatch} and \emph{3DLoMatch} datasets.}
	\label{tab:detailed_3dmatch}
\end{table*}

\subsection{Additional results}
\label{sec:additional_results_supp}
\parahead{Detailed registration results}
We report detailed per-scene \textit{Registration Recall (RR)}, \textit{Relative Rotation Error (RRE)} and \textit{Relative Translation Error (RTE)} in Tab.~\ref{tab:detailed_3dmatch}. RRE and RTE are only averaged over successfully registered pairs for each scene, such that the numbers are mot dominated by gross errors from complete registration failures. We get the highest RR and lowest or second lowest RTE and RRE for almost all scenes, this further shows that our overlap attention module together with probabilistic sampling supports not only robust, but also accurate registration.

\begin{figure*}[t!]
    \centering
    \includegraphics[width=\linewidth]{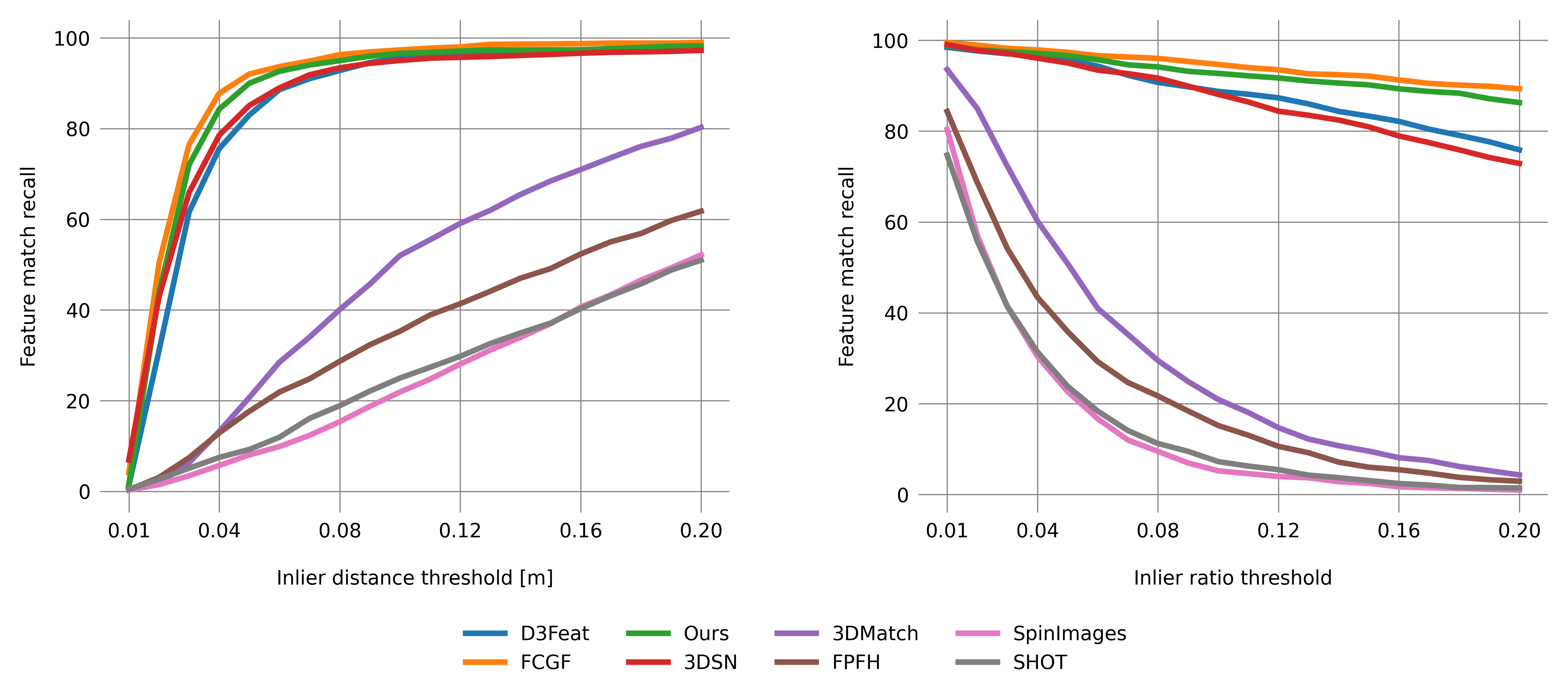}\vspace{-2mm}

    \caption{Feature matching recall in relation to inlier distance threshold $\tau_{1}$ (left) and inlier ratio threshold $\tau_{2}$ (right)}
    \label{fig:fmr}
    \vspace{-\baselineskip}
\end{figure*}

\parahead{Feature match recall}
Finally, Fig.~\ref{fig:fmr} shows that our descriptors are robust and perform well over a wide range of thresholds for the allowable inlier distance and the minimum inlier ratio. Notably, \acro\ consistently outperforms D3Feat that uses a similar KPConv backbone.

\subsection{Additional ablation studies}
\label{sec:addtional_ablation_supp}

\begin{table}[t!]
    \setlength{\tabcolsep}{6pt}
    \renewcommand{\arraystretch}{1.2}
	\centering
	\resizebox{\columnwidth}{!}{
    \begin{tabular}{ll|cccccc}
			\toprule
			\multicolumn{2}{c}{} & \multicolumn{3}{c}{\textit{3DMatch}} & \multicolumn{3}{c}{\textit{3DLoMatch}} \\
            \cline{3-8}
            \scalebox{.8}[1.0]{matchability} & \multicolumn{1}{c}{\scalebox{.8}[1.0]{overlap}} & FMR & IR & RR & FMR & IR & RR \\
            \hline
            \multicolumn{1}{c}{} & \multicolumn{1}{c}{} & \underline{96.2} & 51.6 & 86.0 & 74.9 & 20.4 & 43.3 \\
            \multicolumn{1}{c}{\ding{51}} & \multicolumn{1}{c}{} & 96.1 & 54.0 & \textbf{89.2} & 75.5 & 21.9 & 52.2 \\
            \multicolumn{1}{c}{} & \multicolumn{1}{c}{\ding{51}} & \underline{96.2} & \underline{56.7} & \underline{89.1} & \underline{78.3} & \underline{26.1} & \underline{57.4}\\
            \multicolumn{1}{c}{\ding{51}} & \multicolumn{1}{c}{\ding{51}} & \textbf{96.7} & \textbf{58.0} & 89.0 & \textbf{78.6} & \textbf{26.7} & \textbf{59.8}\\
			\bottomrule
	\end{tabular}
	}
	\caption{Different combinations of scores used for probabilistic sampling.}
	\label{tab:ablate_matchability}
\end{table}
\parahead{Ablations of matchability score}
We find that probabilistic sampling guided by the product of the overlap and matchability scores attains the highest RR. Here we further analyse the impact of each individual component. We first construct a baseline which applies random sampling (\textit{rand}) over conditioned features, then we sample points with probability proportional to overlap scores (\textit{prob. (o)}), to matchability scores (\textit{prob. (m)}), and to the combination of the two scores (\textit{prob. (om)}). As shown in Tab.~\ref{tab:ablate_matchability}, \textit{rand} fares clearly worse, in all metrics. Compared to \textit{prob. (om)}, either \textit{prob. (o)} or \textit{prob. (m)} can achieve comparable results on \emph{3DMatch}; the performance gap becomes big on the more challenging \emph{3DLoMatch} dataset, where our \textit{prob. (om)} is around 4 pp better in terms of RR. 

\begin{table}[t]
    \setlength{\tabcolsep}{6pt}
    \renewcommand{\arraystretch}{1.2}
	\centering
	\resizebox{\columnwidth}{!}{
    \begin{tabular}{lccccc|ccccc}
			\toprule
			& \multicolumn{5}{c|}{\textit{3DMatch}} & \multicolumn{5}{c}{\textit{3DLoMatch}} \\
			\# Samples & 5000 & 2500 & 1000 & 500 & 250 & 5000 & 2500 & 1000 & 500 & 250 \\
            \midrule
			& \multicolumn{10}{c}{\textit{Registration Recall (\%)}} \\
		    \midrule
 			FCGF~\cite{Choy2019FCGF} & 85.1 & 84.7 & 83.3 & 81.6 & 71.4 & 40.1 & 41.7 & 38.2 & 35.4 & 26.8 \\
 			FCGF+OA& \textbf{89.1} & \textbf{88.9} & \textbf{88.7} & \textbf{87.5} & \textbf{85.4} & \textbf{57.8} & \textbf{58.3} & \textbf{59.8} & \textbf{58.7} & \textbf{55.9}\\
			\bottomrule
			
	\end{tabular}
	}
	\caption{Ablation of the proposed overlap attention module with sparse convolution backbone. FCGF + OA denotes adding proposed overlap attention module to FCGF model.}
	\label{tab:ablate_fcgf}
\end{table}

\parahead{Ablations of overlap attention module with FCGF}
To demonstrate the flexibility of our model, we additionally add proposed overlap attention module to FCGF model. We train it on \emph{3DMatch} dataset with our proposed loss for 100 epochs, the results are shown in Tab.~\ref{tab:ablate_fcgf}. It shows that FCGF can also greatly beneﬁt from the overlap attention module. Registration recall almost doubles when sampling only 250 points on the challenging \emph{3DLoMatch} benchmark. 

\subsection{Timings}
\label{sec:timing}
\begin{table}[t!]
    \setlength{\tabcolsep}{6pt}
    \renewcommand{\arraystretch}{1.2}
	\centering
	\resizebox{\columnwidth}{!}{
    \begin{tabular}{ccccccc}
			\toprule
            \multicolumn{2}{c}{} & \multicolumn{1}{c}{\renewcommand{\arraystretch}{0.8}\begin{tabular}{c}\scalebox{0.9}[1.0]{data}\\\scalebox{0.9}[1.0]{loader}\end{tabular}} & \multicolumn{1}{c}{\scalebox{0.9}[1.0]{encoder}} & \multicolumn{1}{c}{\renewcommand{\arraystretch}{0.8}\begin{tabular}{c}\scalebox{0.9}[1.0]{overlap}\\\scalebox{0.9}[1.0]{attention}\end{tabular}} & \multicolumn{1}{c}{\scalebox{0.9}[1.0]{decoder}} &
            \multicolumn{1}{c}{\scalebox{0.9}[1.0]{overall}} \\
            \hline
            \multicolumn{2}{c}{FCGF~\cite{Choy2019FCGF}} & \textcolor{white}{20}\textbf{6} & 414 & --- & 25 & 445 \\
            \multicolumn{2}{c}{D3Feat~\cite{bai2020d3feat}} & 200 & \textcolor{white}{4}\underline{11} & --- & \underline{63} & \underline{274}\\
            \multicolumn{2}{c}{Ours} & \underline{191} & \textcolor{white}{41}\textbf{9} & 70 & \textcolor{white}{6}\textbf{1} & \textbf{271}\\ 
			\bottomrule
	\end{tabular}}
	\caption{Runtime per fragment pair in milli-seconds, averaged over 1623 test pairs of \emph{3DMatch}.}
	\label{tab:3DMatch_times}
\end{table}
We compare the runtime of \acro\ with FCGF%
\footnote{All experiments were done with MinkowskiEngine v0.4.2.} %
~\cite{Choy2019FCGF} and D3Feat%
\footnote{We use its PyTorch implementation.} %
~\cite{bai2020d3feat} on \emph{3DMatch}. For all three methods we set voxel size $V\!=\!2.5\,$cm and batch size 1. The test is run on a single GeForce GTX 1080 Ti with Intel(R) Core(TM) i7-7700K CPU @ 4.20GHz, 32GB RAM. The most time-consuming step of our model, and also of D3Feat, is the data loader, as we have to pre-compute the neighborhood indices before the forward pass. With its smaller encoder and decoder, but the additional overlap attention module, \acro\ is still marginally faster than D3Feat. FCGF has a more efficient data loader that relies on sparse convolution and queries neighbors during the forward pass. See Tab.~\ref{tab:3DMatch_times}.

\subsection{Qualitative visualization}
\label{sec:qualitative_supp}
\begin{figure*}[t]
    \centering
    \includegraphics[width=\textwidth]{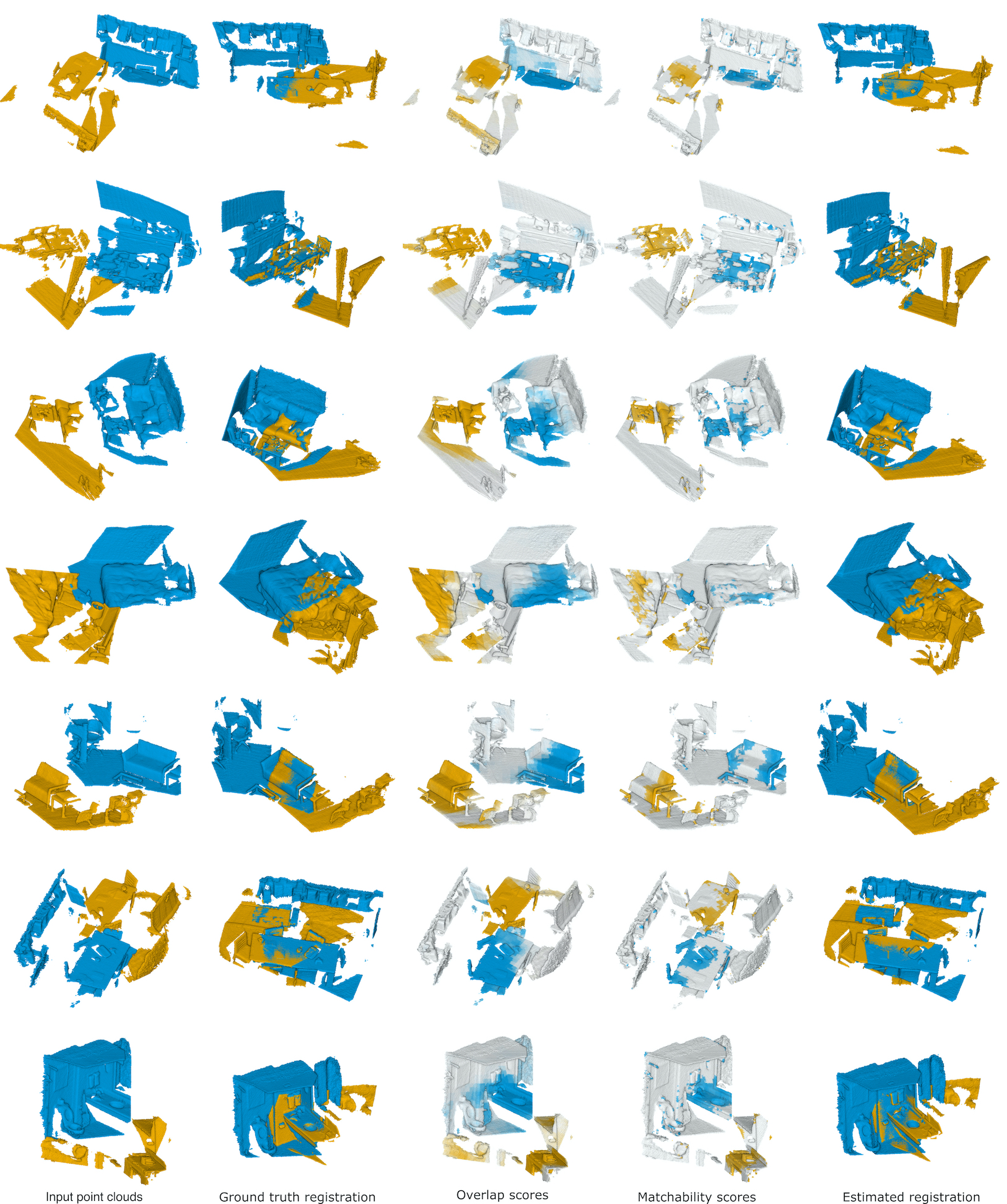}
    \caption{Example results on \emph{3DLoMatch}.}
    \label{fig:3dmatch_supp}
    \vspace{-\baselineskip}
\end{figure*}

\begin{figure*}[t]
    \centering
    \includegraphics[width=0.87\textwidth]{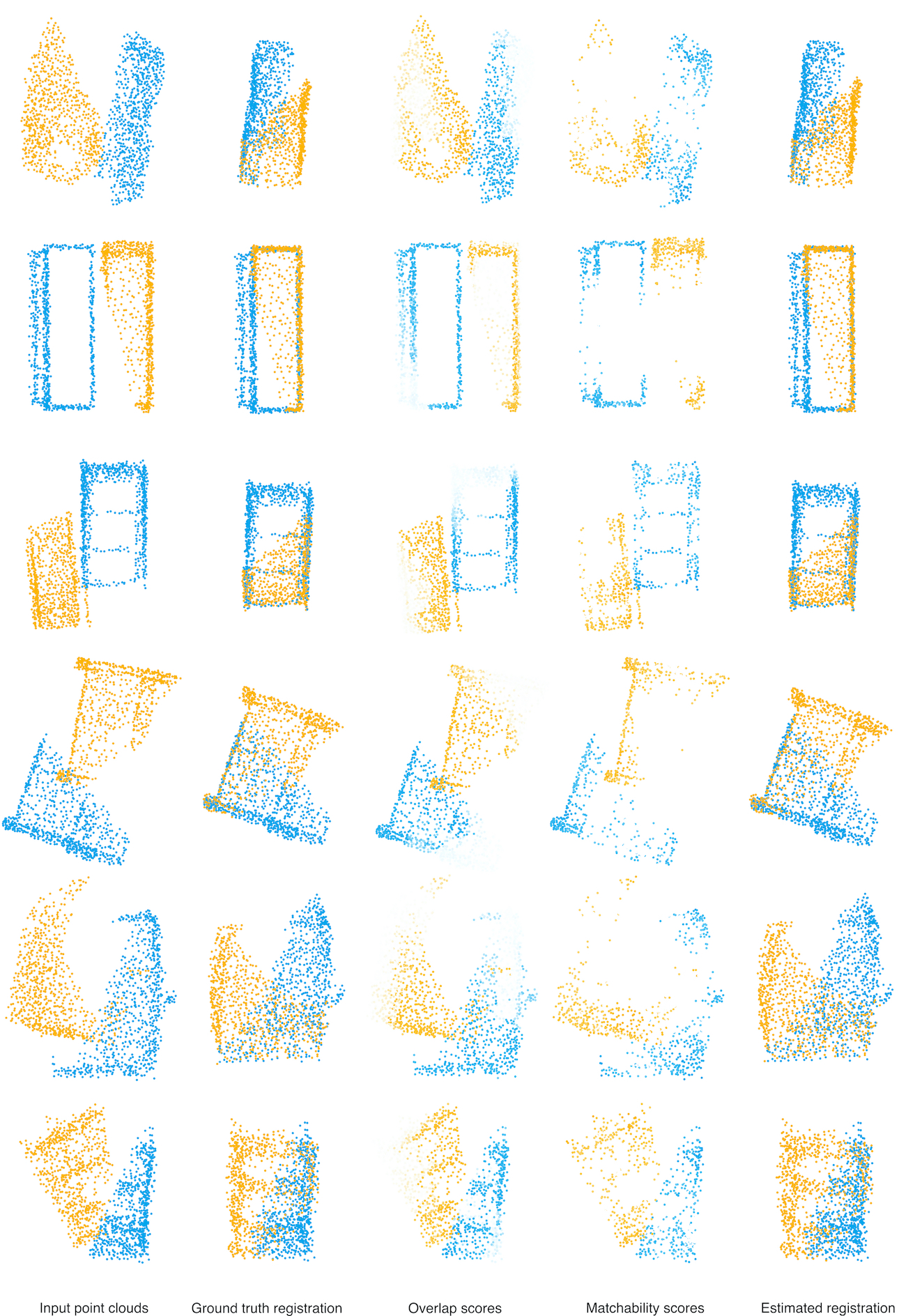}
    \caption{Example results on \emph{ModelLoNet}.}
    \label{fig:modelnet_supp}
    \vspace{-\baselineskip}
\end{figure*}

We show more qualitative results in Fig.~\ref{fig:3dmatch_supp} and Fig.~\ref{fig:modelnet_supp} for \emph{3DLoMatch} and \emph{ModelLoNet} respectively. The input points clouds are rotated and translated here for better visualization of overlap and matchability scores. 


\end{document}